%% file: main_gzh.tex
\pdfoutput=1
\documentclass[11pt]{article}
\input{package_title_author}

\begin{document}
\maketitle

\begin{abstract}

\input{00.Abstract.01}
\end{abstract}

\section{Introduction}
\input{01.Introduction.01}

\section{Related works}
\input{02.RelatedWork.02}

\section{FinED-Bench}
\input{03.Benchmark.01}

\section{Experiments}
\input{04.Experiment.02}

\section{Conclusion}
\input{05.Conclusion.01}

\section*{Limitations}
While FinED-Bench provides a structured evaluation framework, it does not fully capture the complexity of real-world financial scenarios. Several key limitations remain: 1) Diversity of Financial Documents: Many financial documents, such as balance sheets, income statements, are not yet covered. Errors in these documents often originate from underlying data sources, and verifying their correctness often requires a thorough review of extensive historical data. Therefore, we excluded them from the current benchmark.
2) Multimodal Elements: Real-world financial documents often contain visual elements, such as seals and signatures. Accurately interpreting and validating these components requires multimodal capabilities, which are beyond the scope of text-only models evaluated in this paper. Therefore, such types of errors are not considered in the current benchmark.

\section*{Ethical Concerns}
Considering that FinED-Bench may contain sensitive information, such as contact details, even they are publicly available, supervised fine-tuning LLMs on such data could inadvertently amplify security vulnerabilities. To mitigate ethical dilemmas associated with this benchmark, we have invested significant effort and resources to replace real data with carefully crafted synthetic alternatives.

\bibliography{custom}

\newpage

\appendix

\section{LLM Baseline Models Specifications}
\label{app:llm-baselines}

We experiment with a number of eminent general-purpose and domain-specific models from both the API-based and the open-source domains:

\begin{itemize}
    \item \textbf{Qwen3-8B/14B}: the latest generation in the Qwen series, offering a comprehensive suite of dense and MoE models~\cite{qwen3technicalreport}. 
    \item \textbf{DeepSeek-R1-0528-Qwen3-8B}: a distilled model obtained by post-training Qwen3-8B using chain-of-thought traces from DeepSeek-R1-0528~\cite{deepseekai2025deepseekr1incentivizingreasoningcapability}. 
    \item \textbf{Qwen2.5-7B-Instruct}: an improved version of Qwen2, with enhanced capabilities in instruction following, long-text generation, and structured outputs (e.g., JSON)~\cite{qwen2.5}.
    \item \textbf{Dianjin-R1-7B}: a financial domain model built on Qwen2.5-7B-Instruct, incorporating reasoning-augmented supervision and reinforcement learning to enhance financial reasoning~\footnote{https://modelscope.cn/models/tongyi\_dianjin/DianJin-R1-7B}.
    \item \textbf{Fin-R1}: a specialized LLM for financial reasoning, also based on Qwen2.5-7B-Instruct~\footnote{https://modelscope.cn/models/AI-ModelScope/Fin-R1}.
    \item \textbf{GPT-4o-mini} ($\sim$8B parameters): a fast, affordable small model for focused tasks~\cite{gpt4omini}.
    \item \textbf{GPT-3.5-turbo} ($\sim$175B parameters): a ``high-intelligence'' model~\cite{gpt35}.
    \item  \textbf{GPT-4o} ($\sim$200B parameters): a versatile, high-intelligence flagship model~\cite{gpt4omini}.
\end{itemize}

\textbf{Note on Parameter Counts:} The exact numbers of parameters for several LLMs (e.g., GPT series) have not been publicly disclosed yet. The model size estimates reported here are mined from public articles~\footnote{https://www.thealgorithmicbridge.com/p/openai-o1-a-new-paradigm-for-ai}. 

\textbf{Experimental Timeline:} The experimentation was conducted using the official APIs for GPT series between May 11 and May 25, 2025.

The parameter settings used for inference across different LLMs are presented in Table~\ref{tab:pas}.

\begin{table*}[!htbp]
    \centering
    \small
    \begin{tabular}{l|l|l|l|l|l}
    \toprule
        \textbf{Models} & \textbf{Max Tokens} & \textbf{Context Window} & \textbf{Temperature }& \textbf{TopP} & \textbf{TopK }  \\ \midrule
        Qwen2.5-7B-Instruct & 120,000 & 32,768 & 0.7 & 0.8 & 20 \\
        Qwen3-8B (no thinking) & 120,000& 32,768  & 0.6 & 0.95 & 20 \\
        Qwen3-8B & 120,000 & 32,768 & 0.7 & 0.8 & 20 \\
        DeepSeek-R1-0528-Qwen3-8B & 120,000 & 32,768 &  0.6 & 0.95 & 20 \\
        Qwen3-14B (no thinking) & 120,000 & 32,768 & 0.6 & 0.95 & 20 \\
        Qwen3-14B & 120,000 & 32,768 & 0.7 & 0.8 & 20 \\
        GPT-3.5-turbo & 18,000 &16,384 & 0.9 & - & - \\
        GPT-4o-mini & 18,000 & 200,000 & 0.9 & -  & - \\
        GPT-4o & 18,000 & 128,000 & 0.9 & -  & - \\
        Fin-R1 & 120,000 & 32,768 &  0.7 & 0.8 & 20 \\
        Dianjin-R1-7B &  120,000 & 32,768 & 0.7 & 0.8& 20 \\
        \bottomrule
    \end{tabular}
    \caption{Parameter Setting for the inference stage.}
    \label{tab:pas}
\end{table*}

\section{Definitions of Different Errors}
\label{sec:appa}
In the real world, the types of errors found in financial documents are diverse and virtually limitless. Drawing on insights from financial experts and the cognitive-linguistic theory, FinED-Bench focuses on 15 representative subcategories of errors. Below, we will provide specific definitions and examples for each.

\subsection{General Knowledge Errors}
Common mistakes that violate general knowledge include the following five types:
\begin{itemize}
    \item Illegal Time: Dates or times mentioned in the document are inconsistent with known facts.
        
        \begin{tcolorbox}[colback=gray!5, colframe=black, title=Examples]
        {\small \textcolor{red}{\ding{55}}: ... On \textcolor{red}{April 31}, Hangzhou initiated the fourth round of land supply.

        \textcolor{deepgreen}{\ding{52}}:  ...On \textcolor{deepgreen}{April 30}, Hangzhou initiated the fourth round of land supply.}
        \end{tcolorbox}
    \item Redundant Statements: Repeating the same information or including unnecessary repetition in the document.
        
       \begin{tcolorbox}[colback=gray!5, colframe=black, title=Examples]             
       {\small \textcolor{red}{\ding{55}}: ...construction control area: 84,979 square meters, \textcolor{red}{floor area ratio: 2.8}, height limit: 60 meters, \textcolor{red}{floor area ratio: 2.8}...
        
        \textcolor{deepgreen}{\ding{52}}: ...construction control area: 84,979 square meters, \textcolor{deepgreen}{floor area ratio: 2.8}, height limit: 60 meters...}
        \end{tcolorbox}
        
    \item Value Format Errors: The format of attribute values (such as phone numbers, dates, etc.) does not meet standard specifications or expectations.
        
       \begin{tcolorbox}[colback=gray!5, colframe=black, title=Examples]         {\small \textcolor{red}{\ding{55}}: ...Contact number: \textcolor{red}{010}.

        \textcolor{deepgreen}{\ding{52}}: ...Contact number: \textcolor{deepgreen}{010-57365240}.}
        \end{tcolorbox}

    \item Numerical Missing: This error refers to the absence of a required numerical figure in a financial document. 

       \begin{tcolorbox}[colback=gray!5, colframe=black, title=Examples]         {\small \textcolor{red}{\ding{55}}: The land transfer area of plot JG0404-11 in Dingqiao Unit is \textcolor{red}{square meters}.
       
        \textcolor{deepgreen}{\ding{52}}: The land transfer area of plot JG0404-11 in Dingqiao Unit is \textcolor{deepgreen}{30,514 square meters}.}
        \end{tcolorbox}
    
    \item Non-Numerical Attribute Value Missing: This error occurs when a non-numerical attribute (such as names, categories, etc.) misses its corresponding value.
        
       \begin{tcolorbox}[colback=gray!5, colframe=black, title=Examples]         {\small  \textcolor{red}{\ding{55}}: (1) \textcolor{red}{Procurement project name: ;} (2)...

        \textcolor{deepgreen}{\ding{52}}: (1) Procurement project name: \textcolor{deepgreen}{Office Software Procurement}; (2)...}
        \end{tcolorbox}

\end{itemize}
\subsection{Financial Domain Knowledge Errors}
\begin{itemize}
    \item Terminology Misuse: Inaccurate or inappropriate use of financial or industry-specific terms, resulting in incorrect or misleading expressions.
    
       \begin{tcolorbox}[colback=gray!5, colframe=black, title=Examples]         {\small \textcolor{red}{\ding{55}}: In 2024, the company obtained substantial equity funding through \textcolor{red}{debt financing}, thus strengthening its capital structure.
        
        \textcolor{deepgreen}{\ding{52}}: In 2024, the company obtained substantial equity funding through \textcolor{deepgreen}{equity financing}, thus strengthening its capital structure.}
        \end{tcolorbox}
    
    \item Incorrect Legal Reference: Incorrect references to clauses, regulations, or legal documents.
    
       \begin{tcolorbox}[colback=gray!5, colframe=black, title=Examples]         {\small \textcolor{red}{\ding{55}}: The term ``basic medical insurance'' in this contract refers to the basic medical insurance stipulated in the \textcolor{red}{Regulations on Government Investment of the People's Republic of China}.

        \textcolor{deepgreen}{\ding{52}}: The term ``basic medical insurance'' in this contract refers to the basic medical insurance stipulated in the \textcolor{deepgreen}{Social Insurance Law of the People's Republic of China}.}
        \end{tcolorbox}

    \item Ambiguous Expression: The use of vague or unclear language that may lead to multiple interpretations.

      \begin{tcolorbox}[colback=gray!5, colframe=black, title=Examples]         {\small  \textcolor{red}{\ding{55}}: The purchaser shall not impose any unreasonable conditions on the winning bidder as prerequisites for contract signing, \textcolor{red}{except in special circumstances}.

        \textcolor{deepgreen}{\ding{52}}: The purchaser shall not impose any unreasonable conditions on the winning bidder as prerequisites for contract signing.}
        \end{tcolorbox}

    \item Numerical Unit Error: Use of incorrect or non-standard units for numerical values.

       \begin{tcolorbox}[colback=gray!5, colframe=black, title=Examples]         {\small \textcolor{red}{\ding{55}}: Centralized drinking water source project... planned duration: \textcolor{red}{180 hours}.
        
        \textcolor{deepgreen}{\ding{52}}: Centralized drinking water source project... planned duration: \textcolor{deepgreen}{180 calendar days}.}
        \end{tcolorbox}

    \item Omitted Financial Element: Omission of commonly required financial elements, such as bidder qualification criteria, comparative growth rates, etc.
        
       \begin{tcolorbox}[colback=gray!5, colframe=black, title=Examples]         {\small \textcolor{red}{\ding{55}}: In terms of pricing, the price of Grade I metallurgical coke at major ports was RMB 2,540/ton.
        
        \textcolor{deepgreen}{\ding{52}}: In terms of pricing, the price of Grade I metallurgical coke at major ports was RMB 2,540/ton, \textcolor{deepgreen}{down 3.79\% week-over-week}.}
        \end{tcolorbox}

\end{itemize}
\subsection{Financial Reasoning Errors}
\begin{itemize}
    \item Conflicting Expression: Statements in a document that contradict with each other in meaning or logic.

       \begin{tcolorbox}[colback=gray!5, colframe=black, title=Examples]         {\small \textcolor{red}{\ding{55}}: The project is \textcolor{red}{not eligible for bidding}, and is \textcolor{red}{now open for public bidding}.

        \textcolor{deepgreen}{\ding{52}}: The project is \textcolor{deepgreen}{eligible for bidding}, and is now open for public bidding.}
        \end{tcolorbox}

    \item Time Contradiction: Inconsistent or logically conflicting time-related information within a document.

       \begin{tcolorbox}[colback=gray!5, colframe=black, title=Examples]         {\small \textcolor{red}{\ding{55}}: \textcolor{red}{Submission deadline: May 23}... \textcolor{red}{Bid opening: May 22}...

        \textcolor{deepgreen}{\ding{52}}: Submission deadline: \textcolor{deepgreen}{May 22}... Bid opening: May 22...}
        \end{tcolorbox}

    \item Numerical Inconsistency: Discrepancies in numerical values cited in different sections of the same document.

      \begin{tcolorbox}[colback=gray!5, colframe=black, title=Examples]         {\small  \textcolor{red}{\ding{55}}: \textcolor{red}{Tax rate: 6\%}... \textcolor{red}{Tax rate: 8\%}

        \textcolor{deepgreen}{\ding{52}}: Tax rate: 6\%... \textcolor{deepgreen}{Tax rate: 6\%}}
        \end{tcolorbox}

    \item Calculation Error: Incorrect numerical computations or total values.

       \begin{tcolorbox}[colback=gray!5, colframe=black, title=Examples]         {\small \textcolor{red}{\ding{55}}: Annual demand for hip joint systems: \textcolor{red}{285,995 units (Ceramic-Ceramic: 102,264; Ceramic-Polyethylene: 173,303; Alloy-Polyethylene: 1,042)}.

        \textcolor{deepgreen}{\ding{52}}: Annual demand for hip joint systems: \textcolor{deepgreen}{285,995 units (Ceramic-Ceramic: 102,264; Ceramic-Polyethylene: 173,303; Alloy-Polyethylene: 10,428)}.}
        \end{tcolorbox}

    \item Clause Conflict : Conflicting stipulations across different clauses or sections of the document.

       \begin{tcolorbox}[colback=gray!5, colframe=black, title=Examples]         {\small \textcolor{red}{\ding{55}}: Article 10... \textcolor{red}{Under no circumstances shall the contract be terminated early}... Article 11... \textcolor{red}{The contract may be terminated early}.

        \textcolor{deepgreen}{\ding{52}}: Article 10... Under no circumstances shall the contract be terminated early... \textcolor{deepgreen}{[Remove Article 11]}}
        \end{tcolorbox}

\end{itemize}

\begin{figure*} 
    \centering
    \includegraphics[width=\textwidth]{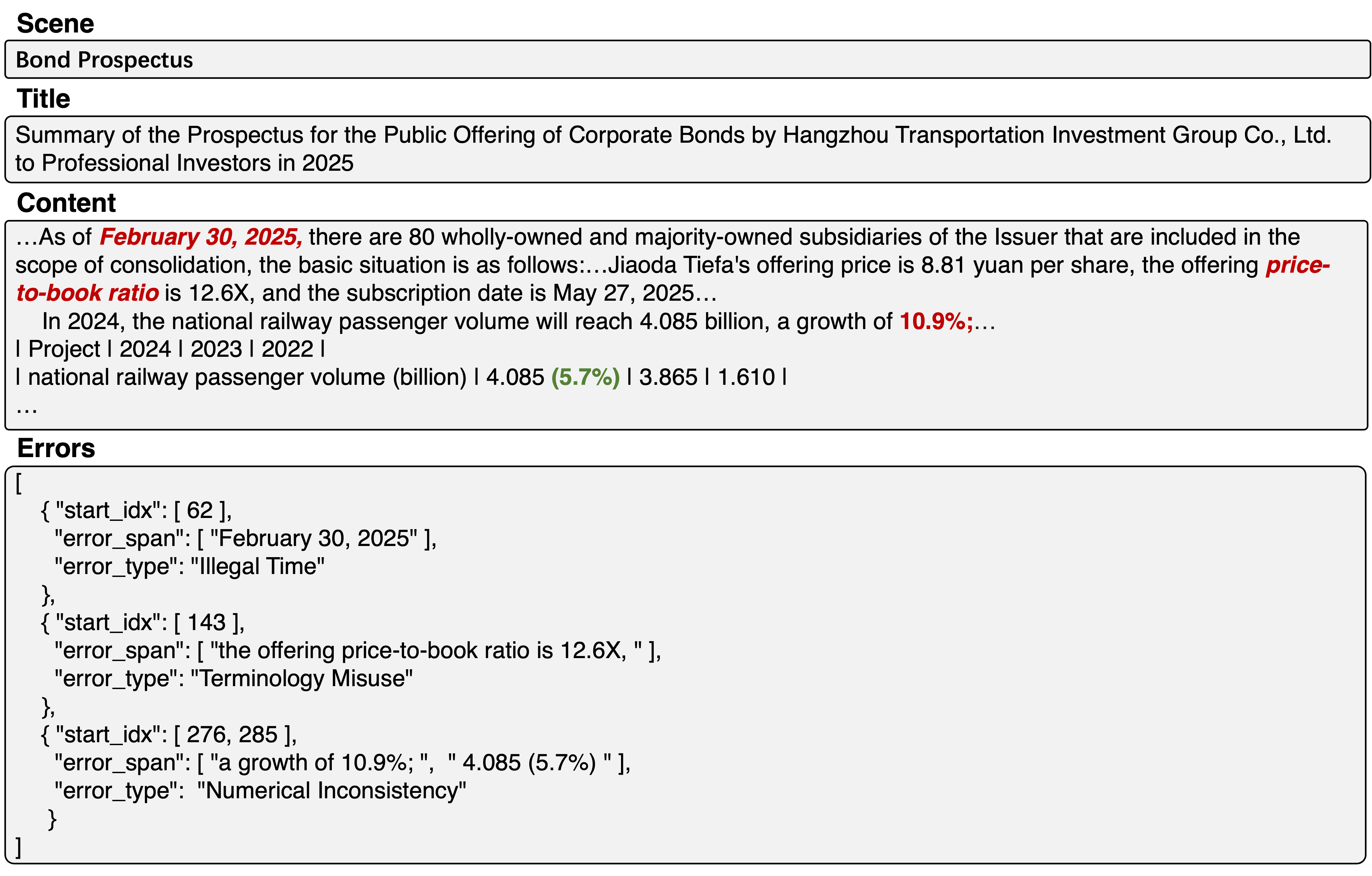}
    \caption{An Example in FinED-Bench.}
    \label{fig:example}
    \vspace{-5mm}
\end{figure*}

An example document in FinED-Bench is shown in Figure~\ref{fig:example}.

\section{Details about Manual Annotation}
\label{sec:anno}
\subsection{Details about Annotators}
Five financial experts served as annotators for this paper. The entire annotation process was conducted under stringent supervision and scrutiny of the first author of this paper.

\begin{table*}
    \centering
    \footnotesize
    \begin{tabular}{l|l}
    \toprule
        \textbf{Task} &  \textbf{Requirements} \\ \hline
        Error verification & \makecell[l]{ 1. Delete if the sentence is actually correct. \\ 2. Delete if its addition would introduce new errors, \\ \quad for example, injecting numerical unit errors may introduce numerical inconsistency.\\ 3. Delete if its incorrectness cannot be inferred from the context. \\ 4. Ensure that each document contains no more than four errors.} \\ \hline
        Seeds Update & \makecell[l]{ 1. Standardize and simplify the definitions of errors. \\ 2. Update the examples of each error type so that LLMs can generate more qualified error instances.} \\ \bottomrule
    \end{tabular}
    \caption{Annotation Requirements for Each Tasks.}
    \label{tab:anno}
\end{table*}

\subsection{Annotation Tasks and Goals}
The purpose of the manual annotation tasks was twofold. The first goal was to obtain a comprehensive annotated dataset that could be used for model evaluation.
The second goal was to modify the definitions of errors and error seeds, enabling LLMs to generate error instances that better reflect real-world scenarios and reducing human participation in the data construction process. All the detailed annotation tasks and targets are list in Table~\ref{tab:anno}.

\subsection{Annotation Consistency}
To ensure the quality of our benchmark dataset, we adopted a majority voting mechanism among five annotators is adopted, with each generated error instance reviewed by three annotators. A high inter-annotator agreement (Fleiss’ $\kappa = 0.89$) indicates that the annotations are consistent and of high quality.

\subsection{Human Performance on the FinED-Bench}
\label{app:human}
In addition to evaluating model performance, we conduct a human baseline study involving two sophomore students majoring in finance on FinED-Bench. Owing to time and cost constraints, we randomly sample 100 documents for manual evaluation and report their average performance in Table~\ref{tab:main_result}. The results indicate that even for human with relevant domain background, accurately identifying all errors in long financial documents remains highly challenging.

\section{Details of Supervised Fine-tuning Data}
\label{sec:ft}
To improve the performance of weaker LLMs in the task of financial error detection, we construct a supervised fine-tuning dataset following the same pipeline as our benchmark, but without human verification. Specifically, after injecting errors, we employ two models (i.e., Qwen3-32B and GPT-4o) as judges to identify both the error type and the error span within the target fragment. An error instance is retained only if both judges correctly detect its span and type. Besides, we preserve the reasoning process generated by Qwen3-32B, located between the <think> and </think> tags. 

The resulting dataset contains 9,515 error-free fragments and 8,697 error-containing fragments, each annotated with the error type, error span and corresponding reasoning process. This dataset is used to fine-tune the target LLM, enabling it to acquire relevant knowledge and improve detection capabilities.

\begin{figure}[!htbp]
    \centering
    \includegraphics[width=0.9\columnwidth]{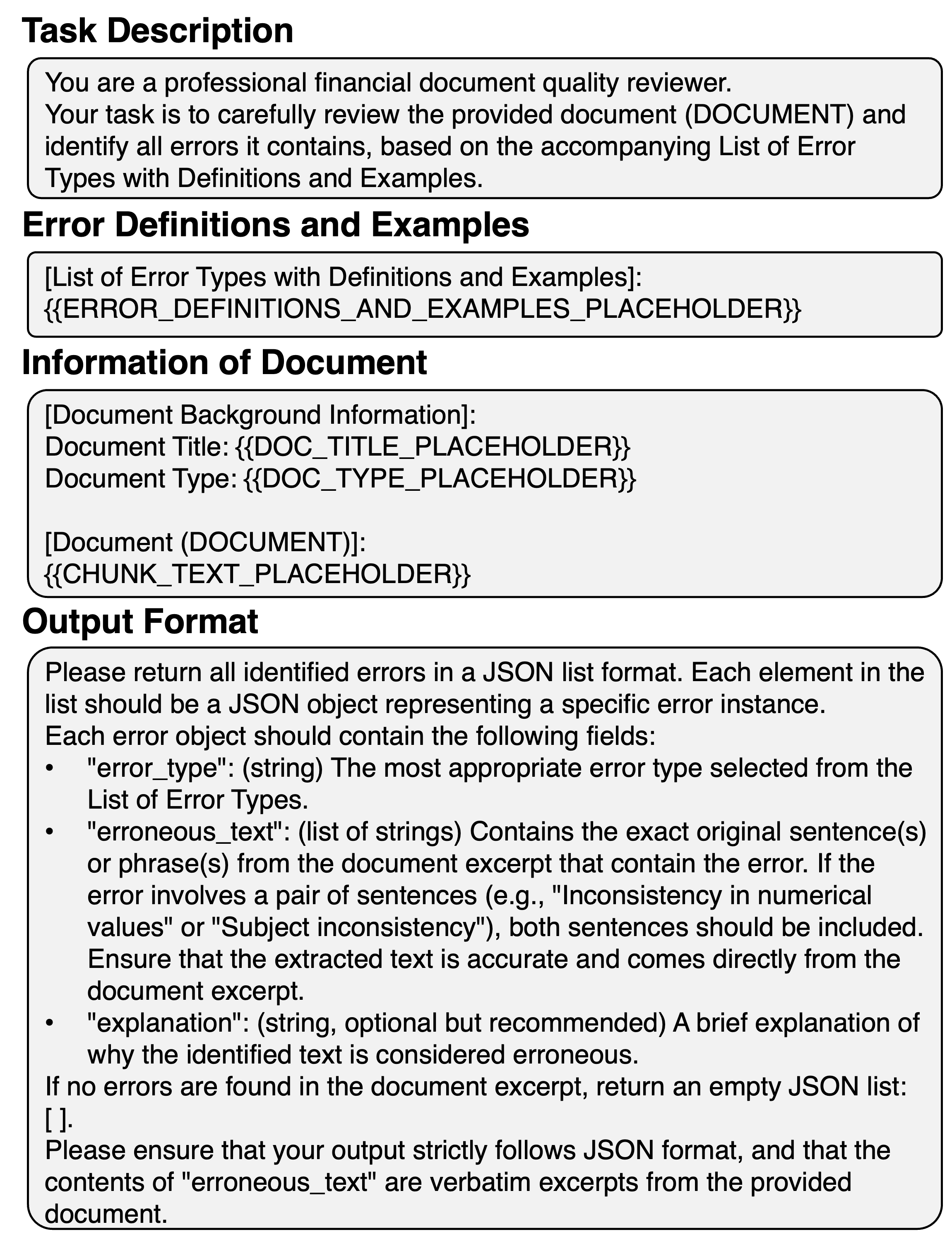}
    \caption{The prompt that guides LLM to perform error detection in financial documents.}
    \label{fig:prompt}
\end{figure}

\begin{figure}[!htbp]
    \centering
    \includegraphics[width=0.9\columnwidth]{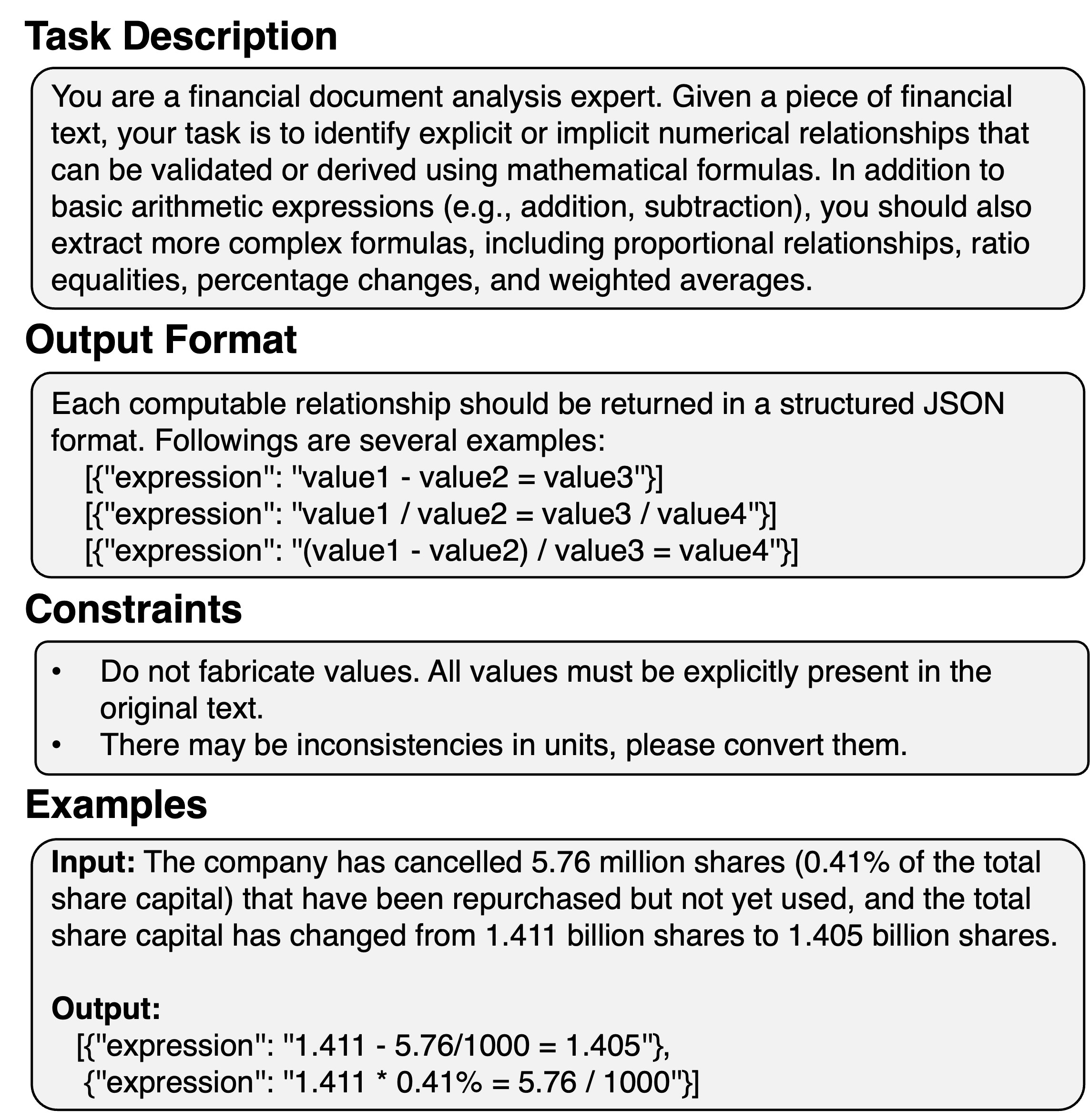}
    \caption{The example of the prompt to find mathematical formulas within a text.}
    \label{fig:cal}
\end{figure}

\section{Prompt Details}
\label{sec:prompt}
This section primarily showcases two prompts used for the evaluation (Figure~\ref{fig:prompt}) and error generation (Figure~\ref{fig:cal}).

\section{Additional Experimental Results}
\label{sec:addex}
\subsection{Prompting Strategies}

\begin{table*}
    \centering
    \small 
    \begin{tabular}{c|c|cccc|cccc}
    \toprule
     \multirow{3}{*}{\textbf{Strategies}} & \multirow{3}{*}{\textbf{Metrics}} & \multicolumn{4}{c|}{\textbf{whole}} & \multicolumn{4}{c}{\textbf{chunk}} \\ \cmidrule(lr){3-10} 
     & & \multicolumn{2}{c|}{\textbf{multi}}& \multicolumn{2}{c|}{\textbf{single}} & \multicolumn{2}{c|}{\textbf{multi}}& \multicolumn{2}{c}{\textbf{single}}  \\ \cmidrule(lr){3-10} 
     &  & \textbf{few} & \textbf{zero} & \textbf{few} & \textbf{zero} & \textbf{few} & \textbf{zero} & \textbf{few} & \textbf{zero} \\ \midrule
\multirow{3}{*}{\makecell{Qwen3-8B \\ (no thinking)}} & Pre. & \textbf{16.87} & 15.70 & 8.71 & 7.39 & 8.00 & 6.57 & 13.30 & 12.78 \\ 
& Rec. & 20.49 & 19.35 & 33.82 & \textbf{36.10} & 33.98 & 35.93 & 21.30 & 18.86 \\ 
& F1 & \textbf{18.50} & 17.33 & 13.85 & 12.27 & 12.95 & 11.11 & 16.38 & 15.24 \\ \midrule
\multirow{3}{*}{Qwen3-8B} & Pre. & \textbf{46.94} & 39.07 & 19.58 & 13.25 & 17.19 & 11.98 & 39.12 & 35.83 \\ 
& Rec. & 34.96 & 33.98 & 42.76 & 43.90 & 43.74 & \textbf{45.20} & 34.80 & 35.77 \\ 
& F1 & \textbf{40.07} & 36.35 & 26.86 & 20.36 & 24.68 & 18.94 & 36.83 & 35.80 \\ \midrule
\multirow{3}{*}{ \makecell{DeepSeek-R1-0528 \\-Qwen3-8B}} & Pre. & \textbf{53.27} & 44.90 & 19.09 & 13.79 & 18.33 & 14.21 & 48.31 & 43.75 \\ 
& Rec. & 18.54 & 17.89 & 17.72 & 14.96 & 17.89 & 17.40 & 18.54 & \textbf{20.49} \\ 
& F1 & 27.51 & 25.59 & 18.38 & 14.35 & 18.11 & 15.64 & 26.80 & \textbf{27.91} \\ \midrule
\multirow{3}{*}{\makecell{Qwen3-14B \\ (no thinking)}} & Pre. & 21.35 & \textbf{21.40} & 11.12 & 10.42 & 10.71 & 9.73 & 17.32 & 18.16 \\ 
& Rec. & 28.78 & 29.27 & 39.51 & 43.25 & 40.16 & \textbf{43.41} & 31.54 & 29.59 \\ 
& F1 & 24.51 & \textbf{24.72} & 17.36 & 16.79 & 16.91 & 15.90 & 22.36 & 22.51 \\ \midrule
\multirow{3}{*}{Qwen3-14B} & Pre. & \textbf{46.04} & 41.20 & 20.63 & 15.50 & 18.97 & 13.16 & 43.15 & 34.18 \\ 
& Rec. & 39.67 & 39.19 & 49.76 & 53.50 & 53.98 & 53.82 & 40.98 & 39.19 \\ 
& F1 & \textbf{42.62} & 40.17 & 29.17 & 24.04 & 28.07 & 21.15 & 42.04 & 36.51 \\ \midrule
\multirow{3}{*}{Avg.} & Pre. & \textbf{36.89} & 32.45 & 15.83 & 12.07 & 14.64 & 11.13 & 32.24 & 28.94 \\ 
& Rec. & 28.49 & 27.94 & 36.71 & 38.34 & 37.95 & \textbf{39.15} & 29.43 & 28.78 \\ 
& F1 & \textbf{30.64} & 28.83 & 21.12 & 17.56 & 20.14 & 16.55 & 28.88 & 27.59 \\
\bottomrule
    \end{tabular}
    \caption{Performance Comparisons under Different Prompt Strategies. \textbf{Bold} indicates the best performance. Note: (1) whole: the entire document is provided in a prompt; (2) chunk: the document is segmented into small chunks and processed sequentially. (3) multi: the LLM is asked to detect all error types at once. (4) single: the LLM is asked to detect one specified error type per query. (5) zero: no in-context examples are provided. (6) few: a small number of examples are given.}
    \label{tab:str4prompt}
\end{table*}

\begin{table*}[!htbp]
    \centering
    \tiny 
    \resizebox{\textwidth}{!}{
    \begin{tabular}{c|c|ccccc|ccccc|ccccc|c}
\toprule
\multirow{2}{*}{\textbf{Models}} & \multirow{2}{*}{\textbf{Metrics}} & \multicolumn{5}{c|}{\textbf{General Knowledge Errors}} & \multicolumn{5}{c|}{\textbf{Financial Domain Knowledge Errors}} & \multicolumn{5}{c|}{\textbf{Financial Reasoning Errors}} & \multirow{2}{*}{\textbf{Overall}} \\ \cline{3-17}
& & IT & RS & VFE & NM & NNM & AE & OFE & UE & TM & ILR & TC & CaE & NI & CE & CC  &\\ \hline
\multirow{3}{*}{Qwen2.5-7B} & Pre. & 31.03 & 18.01 & 21.28 & 25.0 & 12.24 & 0.54 & 2.5 & 36.36 & 4.35 & 2.13 & 18.87 & 6.25 & 3.16 & 3.79 & 18.18 & 14.37 \\ 
 & Rec. & 18.69 & 24.94 & 3.97 & 4.54 & 1.82 & 1.54 & 0.64 & 15.74 & 0.87 & 1.25 & 3.94 & 0.6 & 6.49 & 1.22 & 5.63 & 7.49 \\ 
 & F1 & 23.33 & 20.92 & 6.69 & 7.68 & 3.17 & 0.8 & 1.02 & 21.97 & 1.45 & 1.57 & 6.51 & 1.09 & 4.25 & 1.85 & 8.6 & 9.85 \\ \hline
\multirow{3}{*}{\makecell{Qwen3-8B \\ (no thinking)}} & Pre. & 57.34 & 59.39 & 11.24 & 34.75 & 9.17 & 0.18 & 8.93 & 41.91 & 12.64 & 31.25 & 31.08 & 0.0 & 3.24 & 9.42 & 26.09 & 16.03 \\ 
 & Rec. & 43.1 & 22.79 & 23.41 & 18.55 & 3.33 & 1.54 & 15.71 & 42.59 & 12.79 & 6.25 & 18.11 & 1.2 & 19.46 & 7.07 & 8.45 & 19.12 \\
  & F1 & 49.21 & 32.94 & 15.19 & 24.19 & 4.89 & 0.33 & 11.38 & 42.25 & 12.72 & 10.42 & 22.89 & 0.0 & 5.56 & 8.08 & 12.77 & 17.44 \\ \hline

\multirow{3}{*}{Qwen3-8B} & Pre. & \textbf{76.89} & \underline{62.11} & \underline{31.67} & 65.57 & \underline{31.39} & 0.0 & \textbf{27.12} & \underline{71.95} & 46.96 & 46.43 & \underline{73.17} & 28.89 & 12.13 & 46.04 & 41.94 & \underline{49.24} \\ 
 & Rec. & 63.10 & 56.05 & 27.78 & 36.2 & 26.06 & 0.0 & 20.51 & 58.80 & 15.70 & \underline{16.25} & \underline{47.24} & 7.78 & 17.84 & 15.61 & \underline{18.31} & 33.66 \\ 
  & F1 & \underline{69.32} & \underline{58.92} & \underline{29.60} & 46.65 & 28.48 & 0.0 & \textbf{23.36} & 64.71 & 23.53 & 24.07 & \textbf{57.42} & \underline{12.26} & 14.44 & 23.32 & 25.49 & 39.99 \\ \hline
  
\multirow{3}{*}{\makecell{Deepseek-R1- \\0528-Qwen3-8B}} & Pre. & \underline{70.87} & \textbf{65.14} & \textbf{41.54} & 58.39 & 28.21 & 0.0 & 17.95 & 66.07 & 52.50 & \underline{75.00} & 63.49 & \underline{35.29} & 12.12 & 45.76 & 41.18 & 49.92 \\ 
 & Rec. & 31.03 & 26.51 & 10.71 & 18.1 & 10.0 & 0.0 & 2.24 & 34.26 & 6.1 & 7.5 & 15.75 & 3.59 & 6.49 & 6.59 & 9.86 & 15.21 \\ 
   & F1 & 43.17 & 37.69 & 17.03 & 27.63 & 14.77 & 0.0 & 3.99 & 45.12 & 10.94 & 13.64 & 25.24 & 0.01 & 8.45 & 11.51 & 15.91 & 23.31 \\ \hline
\multirow{3}{*}{\makecell{Qwen3-14B \\ (no thinking)}} & Pre. & 51.84 & 33.49 & 19.06 & \underline{68.07} & 17.54 & \textbf{23.18} & 16.36 & 46.32 & 30.77 & 3.79 & 27.02 & 0.0 & 10.08 & 20.17 & 8.0 & 25.28 \\ 
 & Rec. & 53.2 & 50.0 & \underline{39.29} & 19.61 & 21.02 & \textbf{30.72} & \underline{30.84} & 46.58 & 8.76 & 6.25 & 29.26 & 1.7 & 17.56 & 17.31 & 11.27 & 29.41 \\ 
   & F1 & 52.51 & 40.11 & 25.66 & 30.45 & 19.13 & \textbf{26.42} & \underline{21.38} & 46.45 & 13.64 & 4.72 & 28.09 & 0.0 & 12.81 & 18.63 & 9.36 & 27.19 \\ \hline
   
\multirow{3}{*}{Qwen3-14B} & Pre. & 69.7 & 56.49 & 31.54 & \textbf{73.77} & \textbf{38.11} & 3.45 & \underline{21.89} & \textbf{78.7} & \textbf{69.23} & 52.27 & \textbf{73.91} & \textbf{42.86} & \textbf{17.55 }& \underline{49.42} & 52.17 & \textbf{50.77} \\ 

 & Rec. & 71.38 & \underline{57.67} & 30.16 & 40.72 & \underline{47.58} & 1.54 & 18.59 & \underline{61.57} & \underline{18.31} & \textbf{28.75} & 33.46 & \underline{12.57} & \underline{23.24} & \underline{20.73} & 16.90 & \underline{37.52} \\ 
   & F1 & \textbf{70.53} & 57.08 & \textbf{30.83} & \underline{52.48} & \textbf{42.32} & 2.13 & 20.1 & \underline{69.09} & \underline{28.97} & \underline{37.10} & \underline{46.07} & \textbf{19.44} & \textbf{20.00} & \underline{29.21} & \underline{25.53} & \underline{43.15} \\ \hline
   
\multirow{3}{*}{GPT-3.5-turbo} & Pre. & 32.86  & 0.00  & 9.38  & 0.00  & 25.00  & 0.00  & 0.00  & 7.32  & 0.00  & \textbf{100.00}  & 0.00  & 0.00  & \underline{16.67}  & 0.00  & 14.29  & 13.14 \\
 & Rec. & 46.94  & 0.00  & 10.91  & 0.00  & 1.75  & 0.00  & 0.00  & 16.22  & 0.00  & 1.25  & 0.00  & 0.00  & 8.82  & 0.00  & 1.41  & 4.93 \\ 
  & F1 & 38.66 & 0.00 & 10.08 & 0.00 & 3.28 & 0.00 & 0.00 & 10.08 & 0.00 & 0.02 & 0.00 & 0.00 & 11.54 & 0.00 & 2.56 & 7.17\\ \hline
  
\multirow{3}{*}{GPT-4o-mini} & Pre. & 43.21  & 11.24  & 15.48  & 27.18  & 30.77  & 0.00  & 0.00  & 23.21  & 0.00  & 0.00  & 23.81  & 0.00  & 7.58  & \textbf{50.00 } & \textbf{75.00} & 21.94   \\ 
 & Rec. & \underline{71.43}  & 18.18  & 23.64  & \underline{45.16}  & 28.07  & 0.00  & 0.00  & 35.14  & 0.00  & 0.00  & 15.38  & 0.00  & 14.71  & 1.03  & 8.45  & 16.61  \\
   & F1 & 53.85  & 13.89  & 18.71  & 33.94  & 29.36  & 0.00  & 0.00  & 27.96  & 0.00  & 0.00  & 18.69  & 0.00  & 10.00  & 2.02  & 15.19  & 18.90   \\ \hline

\multirow{3}{*}{GPT-4o} & Pre. & 43.02 & 45.87 & 17.22 & 54.92 & 28.05 & 0.00 & 19.17 & 73.85 & 60.00 & 60.47 & 38.89 & 6.25 & 9.49 & 31.90 & \underline{56.79} & 37.76 \\
 & Rec. & \textbf{90.69} & \textbf{82.78} & \textbf{49.21} & \textbf{74.44} & \textbf{68.00} & 0.00 & \textbf{48.57} & \textbf{90.00} & \textbf{66.67} & 27.08 & \textbf{80.00} & \textbf{22.22} & \textbf{60.00} & \textbf{67.68} & \textbf{43.40} & \textbf{67.13} \\
 & F1 & 58.36 & \textbf{59.03} & 25.51 & \textbf{63.21} & \underline{39.72} & 0.00 & 27.49 & \textbf{81.13} & \textbf{63.16} & \textbf{37.41} & 52.34 & 9.16 & \underline{16.39} & \textbf{43.37} & \textbf{49.20} & \textbf{48.34} \\ \hline
 
\multirow{3}{*}{Fin-R1} & Pre. & 5.54 & 7.93 & 0.79 & 1.36 & 0.3 & 0.0 & 0.32 & 2.55 & 0.0 & 0.0 & 0.79 & 0.6 & 1.08 & 0.24 & 0.0 & 1.9 \\ 
 & Rec. & 10.56 & 14.45 & 1.52 & 2.64 & 0.6 & 0.0 & 0.64 & 4.96 & 0.0 & 0.0 & 1.56 & 1.18 & 1.98 & 0.47 & 0.0 & 3.19 \\ 
   & F1 & 10.56 & 14.45 & 1.52 & 2.64 & 0.6 & 0.0 & 0.64 & 4.96 & 0.0 & 0.0 & 1.56 & 1.18 & 1.98 & 0.47 & 0.0 & 3.19 \\ \hline
\multirow{3}{*}{Dianjin-R1-7B} & Pre. & 53.96 & 20.29 & 18.94 & 38.67 & 8.55 & \underline{8.33} & 3.2 & 60.98 & 18.92 & 33.33 & 30.0 & \underline{18.75} & 5.65 & 27.03 & 13.33 & 27.25 \\ 
 & Rec. & 25.95 & 19.58 & 9.92 & 13.15 & 3.03 & \underline{1.54} & 1.29 & 28.94 & 2.03 & 2.5 & 15.35 & 1.8 & 3.78 & 2.44 & 2.82 & 11.13 \\ 
   & F1 & 35.05 & 19.93 & 13.02 & 19.63 & 4.47 & \underline{2.6} & 1.83 & 39.25 & 3.67 & 4.65 & 20.31 & 3.28 & 4.53 & 4.47 & 4.65 & 15.81 \\ \bottomrule
    \end{tabular}}
    \caption{Performance Comparison of different Large Language Models in FinED-Bench across 15 errors categories. (\%) \textbf{Bold} indicates the best performance and \underline{underlined} indicates the second-best performance within each metric. }
    \label{tab:main_result}
\end{table*}

\begin{table}
    \centering
    \small
    \begin{tabular}{c|c|c|c}
    \toprule
         \textbf{Model} & \textbf{Beginning} & \textbf{Middle} & \textbf{Ending} \\ \midrule
        \makecell{Qwen3-8B \\(no thinking)} & 7.76 & 6.01 & 9.11 \\ 
        Qwen3-8B & 13.04 & 9.19 & 16.75 \\ 
         \makecell{DeepSeek-R1-0528\\-Qwen3-8B} & 6.52 & 4.95 & 7.23 \\ 
         \makecell{Qwen3-14B \\ (no thinking)} & 9.32 & 12.01 & 16.75 \\ 
        Qwen3-14B & 22.67 & 13.07 & 20.63 \\ 
        GPT-3.5-turbo & 1.55 & 11.06 & 3.56 \\ 
        GPT-4o-mini & 0.00 & 0.00 & 13.70 \\ 
        DianJin-R1-7B & 2.48 & 2.12 & 5.45 \\ 
        Fin-R1 & 0.62 & 0.00 & 0.63 \\ 
        \bottomrule
    \end{tabular}
    \caption{Performance different in potions of errors appear in the document.}
    \label{tab:pos}
\end{table}

To find an effective prompting strategy for detecting errors in long financial documents, we evaluate several prompt designs, with the results summarized in Table~\ref{tab:str4prompt}. The findings reveal three key observations.

\quad (1) \textbf{Whole-document prompting is more effective than chunking-based prompting.} As noted in Section~\ref{sec:gen}, many generated errors are designed to span the entire document to increase the difficulty of the benchmark. Such cross-document inconsistencies cannot be reliably captured when the text is divided into isolated chunks, making whole-document prompts more effective.

\quad (2) \textbf{Multi-error prompting outperforms single-error prompting.} Prompting the model to detect all error types simultaneously (`multi') consistently yields higher F1 scores than prompting for one error type at a time (`single'). This indicates that LLMs benefit from cross-error contextual cues when performing multi-type detection.

\quad (3) \textbf{The combination of `whole + multi + few' prompting achieves the best performance.} This strategy attains the highest average F1 (30.64\%) across all tested models, as it provide complete document context, multiple error types, and informative demonstrations.

\subsection{Fine-grained Performance on 15 Error Subcategories}
We show the performance of all models across 15 subcategories of errors in Table~\ref{tab:main_result}.

 \begin{figure}
    \centering
    \small
    \begin{tikzpicture}
    \begin{axis}[
        ybar,
        bar width=12pt,
        width=0.95\linewidth,
        height=5.5cm,  
        ymin=0,
        ymax=550,      
        ylabel={\normalsize Count},
        xlabel={\normalsize Document Length Range (K=1,000)},
        xtick=data,
        xticklabels={
            {$\leq$2K},
            {2K-4K},
            {4K-6K},
            {6K-8K},
            {8K-10K},
            {10K-12K},
            {12K-14K},
            {14K-16K},
            {>32K-64K},
            {64K-120K}
        },
        x tick label style={
            rotate=35,           
            anchor=east, 
            font= \small, 
            yshift=-2pt,
            xshift=-1pt
        },
        tick label style={font=\large},
        label style={font=\large},
        nodes near coords,
        nodes near coords align={vertical},
        every node near coord/.append style={
            font=\large, 
            yshift=3pt,          
            color=teal!60!black
        },
        enlarge x limits=0.05,   
        axis on top,
        xlabel style={
            font=\large, 
            at={(axis description cs:0.5,-0.28)}, 
            anchor=north
        },
        ylabel style={font=\large},
        major grid style={draw=gray!10},
        grid=both,
    ]
    \addplot+[fill=teal!40!white,draw=teal!80!black] coordinates {
        (0,485)
        (1,220)
        (2,72)
        (3,80)
        (4,43)
        (5,28)
        (6,17)
        (7,29)
        (8,12)
        (9,12)
    };
    \end{axis}
    \end{tikzpicture}
    \caption{Document length distribution of FinED-Bench.}
    \label{fig:my_label}
\end{figure}

\subsection{Position-aware Error Detection Performance}
Besides, we analyze model performance with respect to the positional distribution of errors within a document. Specifically, each document is divided into three segments (i.e., beginning, middle, and ending), as shown in Table~\ref{tab:pos}. The results show that models are more effective at detecting errors located in the ending section, followed by the beginning, while errors in the middle section are the most difficult to identify. This pattern aligns with well-established cognitive phenomena in psychology, namely the Primacy Effect and the Recency Effect. Information presented at the beginning of a document is often repeatedly attended to as the model builds a global understanding of the context, whereas information near the end remains salient due to its proximity to the prediction step. In contrast, content in the middle is less likely to be revisited or emphasized during inference, making errors in this region easier to overlook.

\begin{table}
    \centering
    \small
    \begin{tabular}{cccc}
    \toprule
         \textbf{Models} & \textbf{Pre.} & \textbf{Rec.} & \textbf{F1} \\ \midrule
        Qwen3-8B (no thinking) & 13.65 & 24.67 & 17.58 \\ 
        Qwen3-8B & 39.31 & 38.00 & 38.64 \\ 
        DeepSeek-R1-0528-Qwen3-8B & \underline{54.41} & 24.67 & 33.94 \\ 
        Qwen3-14B (no thinking) & 17.12 & 38.00 & 23.60 \\ 
        Qwen3-14B & 47.37 & \underline{42.00} & \underline{}{44.52} \\ 
        Llama3.1-8B & 12.50 & 13.33 & 12.90 \\ 
        GPT-4o-mini & 22.56 & 20.00 & 21.20 \\ 
        GPT-5 & \textbf{57.25} & \textbf{50.00} & \textbf{53.38} \\ 
        Dianjin-R1-7B & 14.06 & 6.00 & 8.41 \\ 
        \bottomrule
    \end{tabular}
    \caption{Performance of Different LLMs on English dataset (\%). \textbf{Bold} indicates the best performance and \underline{underlined} indicates the second-best performance within each metric.}
    \label{tab:res4en}
\end{table}

\subsection{Evaluation on the English Benchmark}
\label{sec:enben}
To address concerns about generalization beyond Chinese, we additionally construct an English financial error-detection dataset sourced from English-language financial websites. This dataset comprises 56 documents with 198 annotated errors, with 91 general knowledge errors, 37 financial domain knowledge errors and 70 financial reasoning errors. The evaluation results are summarized in Table~\ref{tab:res4en}. Overall, the findings are consistent with those observed on the Chinese benchmark:

\quad (1) \textbf{Reasoning-enabled models substantially outperform their non-reasoning versions.} For example, Qwen3-14B improves from 23.60\% → 44.52\% F1, and Qwen3-8B improves from 17.58\% → 38.64\% F1 once thinking is enabled.
    
\quad (2) \textbf{Recall remains a major challenging across models.} Despite improvements in precision and overall F1, recall values generally remain below 50\%, indicating persistent difficulty in comprehensively identifying all errors.
    
\quad (3) \textbf{Financial-domain LLMs remain constrained by the capacity of their base models}. Although these domain-specific LLMs are fine-tuned for tasks like financial QA, text summarization, and classification tasks, only a limited amount of domain-specific knowledge is learned.

\begin{table*}
    \centering
    \scriptsize
    \begin{tabular}{c|c|c|c|c|c|c|c|c}
        \toprule
         \textbf{Models} & \textbf{Base Model} & \textbf{Paras.} & \textbf{Length} & \textbf{Tasks} & \textbf{Techniques} & \textbf{Chinese}& \textbf{Year} & \textbf{Open} \\ \hline
         Plutus~\cite{peng2025plutus} & Llama & 8B & 42000  & SMP & IFT  &  \ding{55} & 03/03/2025 & \ding{52}\\\hline
         BloomberGPT~\cite{wu2023bloomberggpt} & BLOOM & 50B & 2048 & \makecell{SA, HC, \\ NER, QA} & PT, PE & \ding{55} &   03/30/2023  & \ding{55} \\ \hline
         FinMA~\cite{xie2023pixiu} & Llama & 7B/13B & 4096 & \makecell{SA, HC, NER, \\ QA, SMP} & IFT, PE & \ding{55}   & 06/01/2023 & \ding{52}\\  \hline
         InvestLM~\cite{kong2024large} & Llama & 65B & 4096 & \makecell{SA, HC,  \\ QA, Summ} & \makecell{IFT, PE,  \\ PEFT} & \ding{55} & 09/15/2023& \ding{52} \\ \hline
         FinGPT-v3~\cite{wang2023fingpt} & Llama2 & 7B & 4096 & \makecell{SA, HC, \\ NER, RE} & \makecell{IFT, PE, \\ PEFT} & \ding{55}  & 10/12/2023  & \ding{52}\\ \hline
         DianJin-R1~\cite{zhu2025dianjin} & Qwen2.5 & 7B/32B & 131072  &  FS, IR  & IFT, RL & \ding{52} &  04/23/2025 & \ding{52}\\\hline
         Fin-R1~\cite{liu2025fin} & Qwen2.5 & 7B & 131072  & FC, FS & IFT, RL & \ding{52}  & 03/22/2025 & \ding{52}\\ \bottomrule
         
    \end{tabular}
    \caption{A Summary of FinLLMs. The abbreviations correspond to Para. = Parameters, PT = Pre-Training, PE = Prompt Engineering, IFT= Instruction Fine-Tuning, PEFT = Parameter Efficient Fine-Tuning, RL = Reinforcement Learning; [SA] Sentiment Analysis, [HC] Headline Classification, [NER] Named Entity Recognition, [QA] Question Answering, [SMP] Stock Movement Prediction, [Summ] Text Summarization, [RE] Relation Extraction, [FS] Financial Services, [IR] Investment Research, [FC] Financial Coding.}
    \label{tab:finllms}
\end{table*}

 \begin{table*}
     \centering
     \footnotesize
     \begin{tabular}{c|c|c|c}
            \toprule
          \textbf{Dataset} &  \textbf{Tasks}  & \textbf{Language} & \textbf{Year}  \\ \hline
          FLUE~\cite{shah2022flue} & TA, IE, QA  & English & 10/31/2022 \\ \hline
          PIXIU~\cite{xie2023pixiu} & TA, IE, QA, FO, RM  &  English & 01/08/2023 \\ \hline
          FinanceBench~\cite{islam2023financebench} & QA  & English & 12/20/2023 \\ \hline
          BizBench~\cite{koncel2023bizbench} & QA, IE, TG  & English & 03/12/2024 \\ \hline
          FOMC~\cite{shah-etal-2023-trillion} & TA  & English & 2023 \\ \hline
          FinBen~\cite{xie2024finben} &  TA, IE, QA, FO, RM, DM & English, Chinese & 2024  \\ \hline
          CFBenchmark~\cite{lei2023cfbenchmark} & IE, TA, TG & Chinese & 05/21/2024 \\ \hline
     \end{tabular}
     \caption{Comparison of Different Financial Models. The abbreviations correspond to IE = Information Extraction, TA = Textual Analysis, QA = Question Answering, TG = Text Generation, RM = Risk Management, FO = Forecasting, DM = Decision-Making.}
     \label{tab:ben4fin}
 \end{table*}
 
\begin{table*}
    \centering
    \footnotesize
    \begin{tabular}{c|c|c|c|c|c}
    \toprule
        \textbf{Datasets} &  \textbf{Domain} & \textbf{Language} & \textbf{Avg. Tokens} & \textbf{Error Types} & \textbf{Year}\\ \hline
        CoNLL-2014~\cite{hernandez2014conll} & General & English  & 602.9  & grammatical, syntactic & 05/2014\\ \hline
        BEA-2019~\cite{bryant2019bea} &  General & English & 244.9  & grammatical & 08/02/2019 \\ \hline
        MEDEC~\cite{abacha2024medec} & Medical & English & 126.5 & semantic & 01/02/2025 \\ \hline
        GMEG~\cite{napoles2019enabling} & General & English & 20.7 & grammatical &  04/2019\\\hline
        NLPTEA-2020~\cite{rao2020overview}  & General & Chinese & 35.4 &  grammatical & 12/04/2020 \\ \hline
        MuCGEC~\cite{zhang-etal-2022-mucgec} & General & Chinese & 38.5 & grammatical & 07/10/2022 \\ \hline
        TEC-JL~\cite{koyama2020construction} & General & Japanese & 21.8 & grammatical & 05/11/2020 \\ \hline
        UA-GEC~\cite{syvokon2021ua} & General & Ukrainian & 15.9  & grammatical &  11/08/2022 
        \\ 
        \bottomrule 
    \end{tabular}
    \caption{Comparison of Different Error Detection Benchmarks.}
    \label{tab:errors}
\end{table*}

\section{Supplements to Related Work}
\label{sec:ben}
We list financial models and their targeted tasks in Table~\ref{tab:finllms}, and existing error detection benchmarks in Table~\ref{tab:errors}.

\end{document}

%% file: package_title_author.tex
\usepackage{acl}
\usepackage{times}
\usepackage{latexsym}
\usepackage[T1]{fontenc}
\usepackage[utf8]{inputenc}
\usepackage{microtype}
\usepackage{inconsolata}

\usepackage{graphicx}
\usepackage[table]{xcolor}
\usepackage{array}
\usepackage{colortbl}
\usepackage{multirow}
\usepackage{makecell}
\usepackage{amsmath}
\usepackage{pifont}
\usepackage[most]{tcolorbox}
\usepackage{listings}
\usepackage{pgfplots}
\usepackage{subcaption}
\usepackage{threeparttable}
\usepackage{booktabs}

\pgfplotsset{compat=newest}

\newcolumntype{C}{>{\columncolor{blue!5}}c}
\newcolumntype{P}{>{\columncolor{blue!10}}c}

\definecolor{deepgreen}{rgb}{0.0, 0.39, 0.0}
\definecolor{bgcolor}{RGB}{242, 243, 245}
\definecolor{xhsbg}{RGB}{240, 248, 255}
\definecolor{xhsred}{RGB}{220, 20, 60}

\newtcolorbox{finding}[1]{
  before={\par\noindent},
  after skip=0pt,
  after=\noindent,
  colback=xhsbg!10,
  colframe=xhsred!70,
  title={Finding #1},
  fonttitle=\bfseries
}

\title{Are Large Language Models Reliable Reviewers?\\
A Benchmark for Error Detection in Financial Documents}

\author{
\textbf{Ying He \textsuperscript{1}},
\textbf{Zhouhong Gu \textsuperscript{1}},
\textbf{Zhecheng Hu \textsuperscript{1}},
\textbf{Yubo Zhou \textsuperscript{1}},
\\
\textbf{Hao Shen \textsuperscript{1}},
\textbf{Jiaqing Liang \textsuperscript{2}},
\textbf{Zhaoqian Dai \textsuperscript{3}},
\textbf{Shuguang Ma \textsuperscript{3}},
\\
\textbf{Fei Yu \textsuperscript{3}},
\textbf{Yanghua Xiao \textsuperscript{1\thanks{Corresponding authors.}}},
\textbf{Zhixu Li \textsuperscript{4\footnotemark[1]}}
\\
\textsuperscript{1} College of Computer Science and Artificial Intelligence, Fudan University,
\\
\textsuperscript{2} School of Data Science, Fudan University,
\textsuperscript{3} Ant Group,
\\
\textsuperscript{4} School of Information and School of Smart Governance, Renmin University of China
\\
\texttt{\{yinghe23,zhouyb24,zchu24,hshen22\}@m.fudan.edu.cn},
\\
\texttt{\{zhgu20,liangjiaqing,shawyh\}@fudan.edu.cn}, \texttt{zhixuli@ruc.edu.cn}
\\
\texttt{\{daizhaoqian.dzq,liangxiao.msg\}@antgroup.com}, \texttt{feiyu.fyyu@gmail.com}
}

%% file: 00.Abstract.01.tex
Ensuring the accuracy of financial documents is critical for economic analysis, regulatory compliance, and corporate decision-making.
Several studies have shown that Large Language Models (LLMs) perform well in many financial tasks, such as stock price movements and financial analytics. 
However, a critical task remains unexplored: the ability of LLMs to identify errors in financial documents. In this paper, we introduce \textbf{FinED-Bench}, the first publicly \textbf{Bench}mark for \textbf{Fin}ancial \textbf{E}rror \textbf{D}etection across three levels of cognitive complexity. FinED-Bench covers nine real-world financial scenarios, and includes over 900 documents reported in 2025 that are unseen by existing language models. We detail the benchmark construction process and evaluate several advanced LLMs (e.g., GPT-4o, Qwen3-14B) on this tasks, which requires both financial domain knowledge and reasoning capabilities. Experimental results show that current LLMs still struggle with this task, especially in high-complexity cases. Besides, supervised fine-tuning can significantly improve the performance of weaker LLMs on this task. Our data and code are available at https://github.com/hedyHe/FinED-Bench.

%% file: 01.Introduction.01.tex
Errors in financial documents have serious impacts on economic analysis~\cite{wu2023bloomberggpt}, regulatory compliance~\cite{xie2024finben}, and corporate decision-making~\cite{peng2025plutus}. A notable example is the 2012 ``London Whale'' scandal, where JPMorgan Chase's \$6 billion loss was not caused by market volatility but by an Excel error, highlighting the urgent need for error detection in the financial industry. A survey~\cite{gartner2024} further reveals that 18\% of financial practitioners make errors daily, one-third make errors several times weekly, and 59\% make errors several times monthly. These errors not only may lead to huge economic losses but also affect market confidence and the accuracy of regulatory decisions.

\begin{figure}
    \centering
    \includegraphics[width=\columnwidth]{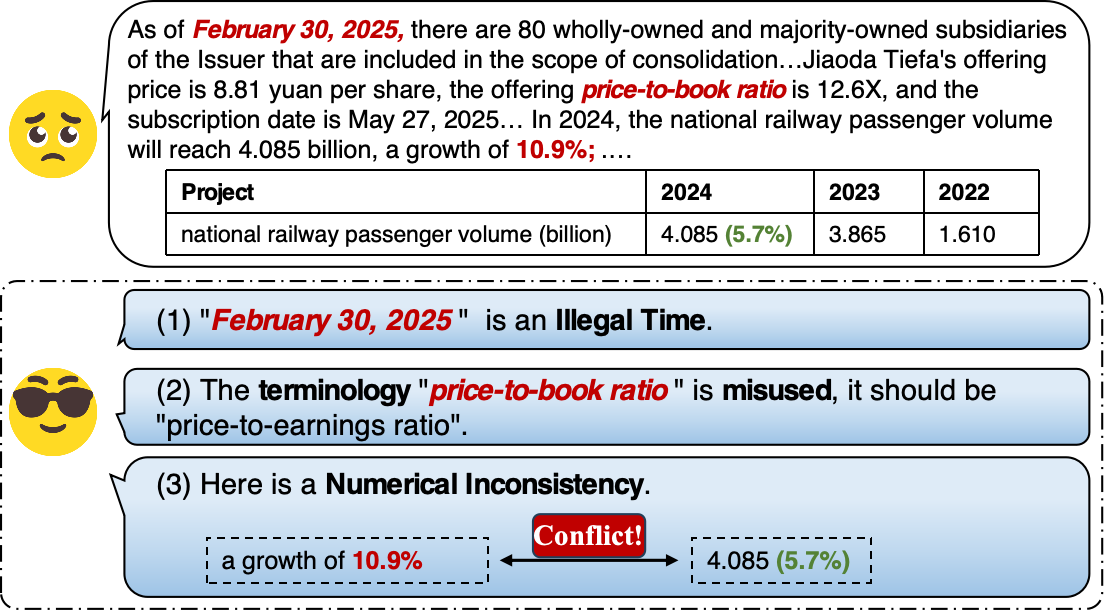}
\caption{An illustrative example of three types of errors in financial documents. The figure demonstrates (1) General Knowledge Errors such as illegal times, (2) Financial Domain Knowledge Errors including terminology misuse, and (3) Financial Reasoning Errors requiring cross-paragraph verification, where passenger volume growth claims in the text contradict numerical data in tables.}
    \label{fig:intro}
\end{figure}

\begin{figure*}[!htbp]
    \centering
    \includegraphics[width=1.0\textwidth]{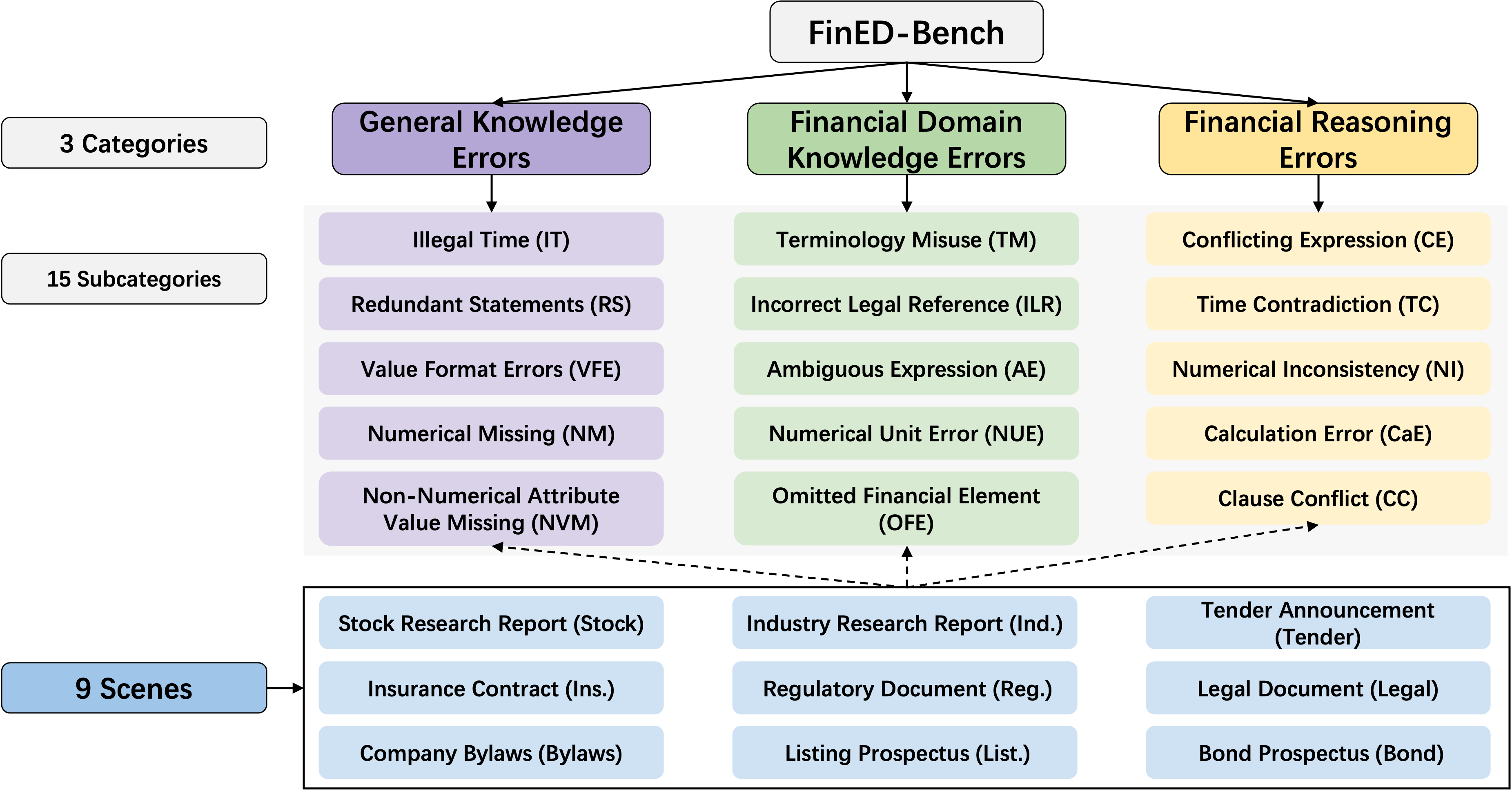}
    \caption{Overview of FinED-Bench error taxonomy and document scenes.
    The benchmark categorizes financial document errors into three hierarchical levels: (1) General Knowledge Errors (GKEs) including format and missing value issues, (2) Financial Domain Knowledge Errors (FKEs) covering terminology and regulatory violations, and (3) Financial Reasoning Errors (FREs) requiring complex inference across document sections. These 15 error subcategories are evaluated across 9 realistic financial scenes ranging from research reports to legal contracts.}

    \label{fig:def-of-errors}
\end{figure*}

Compared to general documents, error detection in financial documents faces unique challenges. 
These errors can be categorized into three levels: general knowledge errors, financial domain-specific knowledge errors, and financial reasoning errors that require complex inference. And, each level comprises several subcategories, as illustrated in Figure~\ref{fig:def-of-errors}. 
For example, general knowledge errors include illegal times like ``February 30, 2025''; domain-specific errors involve terminology misuse such as incorrectly stating ``price-to-earnings ratio'' as ``price-to-book ratio''; reasoning errors manifest as cross-paragraph inconsistencies where a report's front section claims 10.9\% growth but the subsequent table shows 5.7\%, as illustrated in Figure~\ref{fig:intro}. Detecting these errors requires multi-hop reasoning and a firmer grasp of financial domain knowledge compared to general-domain error detection.

Traditional automated error detection methods~\cite{guo2021global,li2022wspeller} mainly rely on rule matching and statistical models, which have obvious limitations when handling semantic understanding, multi-hop reasoning, and complex contextual dependencies. In recent years, Large Language Models (LLMs) have demonstrated remarkable capabilities in natural language understanding, mathematical reasoning, and complex text analysis ~\cite{ahn2024large,nam2024using}, providing new solutions for financial error detection. The strong contextual understanding, cross-domain knowledge integration, and multi-step reasoning capabilities of LLMs make them particularly suitable for handling complex errors in financial documents. However, a critical question arises: \textbf{To what extent can LLMs understand complex financial documents and accurately detect subtle errors within them?}

Although previous benchmarks~\cite{onoe2021creak, abacha2024medec, abacha2024overview} have explored grammatical errors (e.g., ``I goes to school'') and hallucination detection (e.g., inconsistency between a summary and its source paragraph) in generated texts, they typically focus on short, sentence-level texts in general domains. Consequently, they fall short in evaluating the numerical and long-context reasoning capabilities required for financial error detection. The field of professional error detection in financial documents still lacks systematic research and evaluation benchmarks. The market urgently needs a specialized financial error detection benchmark that should have the following characteristics: 1) realistic financial document scenarios; 2) coverage of multi-level error types; 3) evaluation of long document reasoning capabilities.
To fill this gap, we propose the \textbf{Fin}ancial \textbf{E}rror \textbf{D}etection \textbf{Bench}mark (\textbf{FinED-Bench}), the first comprehensive benchmark specifically designed to evaluate LLMs' capabilities in financial document error detection. FinED-Bench contains 973 realistic financial documents (average length: 3,784.6 words) with 4,123 annotated error instances, covering the three levels of error types mentioned above. All error instances are annotated and verified by experienced financial experts, ensuring the accuracy and reliability of the benchmark. To further challenge LLMs' long-context reasoning abilities, we construct the FinED-Bench-Hard subset, which contains 24 documents ranging from 32K to 120K (where K=1,000) words with 83 error instances.

We conduct comprehensive experiments to evaluate various LLMs on FinED-Bench. 
Results reveal that recent competitive LLMs face significant challenges in financial document error detection. 
The best-performing model, GPT-4o, achieves an overall F1 score of 48.34\%, with performance dropping substantially across error categories: from 52.33\% for general knowledge errors to 38.00\% for financial reasoning errors. 
This performance degradation is consistent across all models, highlighting limitations in handling complex financial logic. 
Additionally, document length severely impacts performance, with F1 scores declining from 40.16\% to 16.66\% as document length increases from 2.5K to 50.2K words. 
Notably, financial domain fine-tuning substantially enhances performance, with Qwen3-14B showing a 10.70\% improvement in overall F1 score, demonstrating the importance of domain-specific adaptation for this task.

%% file: 02.RelatedWork.02.tex
\subsection{Financial Evaluation Benchmarks}

LLMs have gained significant attention in the financial domain, being adapted for tasks such as financial text analysis~\cite{zhang2023enhancing, zhang2024finsql}, market sentiment prediction~\cite{delgadillo2024finsosent}, and automated trading strategies~\cite{ding2024large}. Many domain-specific models have been developed, including BloombergGPT~\cite{wu2023bloomberggpt}, FinGPT~\cite{wang2023fingpt}, FinMA~\cite{xie2023pixiu}, OpenFinLLMs~\cite{huang2024open}, and Plutus~\cite{peng2025plutus}. Researchers have also proposed various LLM-based agent systems such as FinAgent~\cite{zhang2024multimodal}, FinMem~\cite{yu2024finmem}, FinCon~\cite{yu2024fincon}, FinVision~\cite{fatemi2024finvision}, and FinRobot~\cite{yang2024finrobot}. Additionally, general-purpose LLMs like Qwen3 series~\cite{qwen3technicalreport} and Deepseek-R1~\cite{deepseekai2025deepseekr1incentivizingreasoningcapability} have shown strong performance on financial tasks. Existing financial benchmarks~\cite{xie2024finben, xie2023pixiu, koncel2023bizbench, sinha2022sentfin, arun2025finreflectkg, tatarinov2025kg} typically evaluate model performance across seven core tasks: Information Extraction (IE), Textual Analysis (TA), Question Answering (QA), Text Generation (TG), Risk Management (RM), Forecasting (FO), and Decision-Making (DM), assessing capabilities in understanding, reasoning, and generation.

Unlike existing financial benchmarks that focus on standard tasks, FinED-Bench specifically targets error detection in lengthy financial documents, providing a unique perspective on model reliability in professional financial contexts.

\subsection{Error Detection Benchmarks}

Errors are prevalent in daily communication, academic writing, and professional workflows, arising from human oversight, data processing issues, or insufficient knowledge. Before the advent of LLMs, the Nature Language Processing (NLP) community primarily focused on formal errors such as spelling, punctuation, grammar, and word choice, which can be detected by rule-based approaches~\cite{bryant2023grammatical,nather2020depth}. Recent work has begun to identify semantic and mathematical errors requiring deeper contextual understanding. Existing benchmarks~\cite{onoe2021creak, abacha2024medec, abacha2024overview} have explored hallucination detection and mathematical errors in generated text. For example, MEDEC~\cite{abacha2024medec} provides a benchmark for detecting and correcting errors in clinical notes, though it is limited to short medical texts with an average length of 126.5 words. ErrorRadar~\cite{yan2024errorradar} introduces a multimodal benchmark for detecting errors in K-12 mathematical problem-solving, but it neglects the real-world relevance of mathematical reasoning in everyday or professional contexts.

To address the gap in document-level financial error detection, FinED-Bench provides the first comprehensive benchmark for detecting semantic errors in lengthy Chinese financial documents, focusing on detection rather than correction across three distinct error levels.

%% file: 03.Benchmark.01.tex
\begin{figure*}[!htbp]
    \centering
    \includegraphics[width=1.0\textwidth]{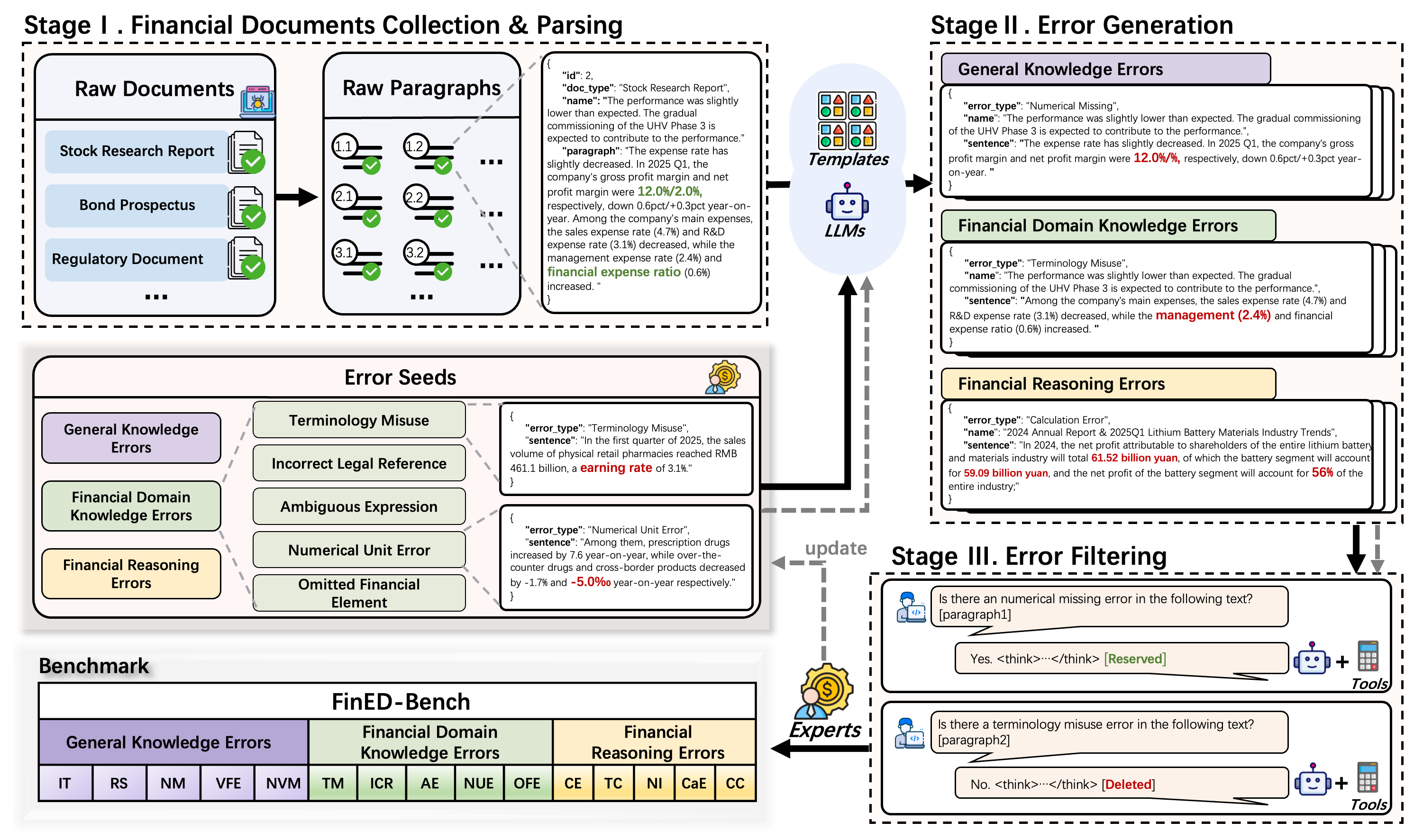}
    \caption{The Semi-Automated Pipeline for Benchmark Construction. The pipeline segments financial documents, injects errors from 15 subcategories (abbreviated as IT, RS, NM, VFE, NVM, TM, ILR, AE, NUE, OFE, CE, TC, NI, CaE, CC), and applies two-stage filtering through model-based verification and manual annotation to ensure benchmark quality.}
    \label{fig:frame4cons}
    \vspace{-3mm}
\end{figure*}

The benchmark is built upon a two-tier error taxonomy which includes 3 categories and 15 subcategories, developed through collaboration between finance domain experts and LLMs.

\subsection{Error Definition}
To ensure systematic and reliable evaluation, we design a two-tier error taxonomy grounded in cognitive-linguistic theory and real-world financial practices. 

Drawing inspiration from the \textit{Discourse Representation Model}~\cite{kintsch1978toward}, which describes three levels of text comprehension, we classify 15 common types of financial errors into the following three major categories:

(1) \textbf{General Knowledge Errors}: Errors at this level affect the readability and surface structure of the text, including redundant statements, incorrect dates, and missing numerical data and so on. Such errors can be detected by non-experts. (2) \textbf{Financial Domain Knowledge Errors}: Errors at this level, such as misuse of terminology, incorrect legal reference, or ambiguous expression, distort the intended meaning. Detecting them typically requires domain-specific knowledge. (3) \textbf{Financial Reasoning Errors}: Logical inconsistencies, such as conflicting expressions, time contradiction, or contradictory clauses, fall into this category. Detecting these errors requires document-level comprehension and reasoning.

\subsection{Dataset Construction}

Given the absence of publicly available erroneous financial documents and concerns regarding training data contamination, we propose a semi-automated construction pipeline in Figure~\ref{fig:frame4cons}.

\subsubsection{Financial Documents Collection \& Parsing}

To avoid overlap with LLM training corpora and minimize the chance of memorization, we collected a set of financial documents published after February 2025, postdating the knowledge cutoffs of evaluated models. To unify the format of the input for LLMs, all documents were converted to plain text. A basic data cleaning process was conducted, including the removal of special characters (e.g., useless HTML tags), blank lines, and other noise using regular expressions. Additionally, documents under 200 words were excluded to ensure substantive content. 

Although FinED-Bench focuses on assessing LLMs’ error detection capabilities in textual data, tabular information, an essential component of financial documents, was also retained to preserve factual integrity. Tables were converted into Markdown format using tools such as OCR or openpyxl, and then integrated into the textual data. Structural errors within tables were manually corrected, while only content-related errors were preserved for evaluation.

\textbf{Hard Set:} Due to the context length limitations of most LLMs (typically 16$k$ or 32$k$ tokens, where $k$=1024), we collect documents below 16K words for the main dataset. However, some financial documents, such as prospectuses, substantially exceed this length yet are important in this domain. We therefore construct a challenging subset, \textbf{FinED-Bench-Hard}, comprising 24 documents ranging from 32K to 120K words, to evaluate model performance under long-context scenarios.   

\subsubsection{Error Generation}
\label{sec:gen}
Combining domain expertise with LLM generation capabilities, we generate errors for each document automatically. The process begins with domain experts identifying and formulating error seeds, which are common instances observed in real-world financial documents. 

Considering the length of many financial documents, especially those exceeding 32K words, injecting errors directly across the entire document poses challenges for LLMs. To address this, we segment documents into shorter, semantically complete fragments. The segmentation strategy varies by error types: (1) For general knowledge and financial domain knowledge errors, documents are divided by chapter, and the obtained fragments serve as candidate contexts. (2) For calculation errors, we select paragraphs with more than three numerical values as the candidates. (3) For other reasoning errors, like time contradiction and numerical inconsistency, we merge adjacent paragraphs sharing overlapping numerical values or terminology as the candidates. To increase the difficulty of the benchmark, we join paragraphs that are originally far apart in the document, for example, merging the first and last paragraphs as a candidate fragment.

This segmentation strategy serves three key purposes: (1) reducing the input length for the generator (GPT-4o), (2) preserving the semantic integrity of candidate fragments, and (3) ensuring an even distribution of error instances across the document.

Following segmentation, we apply In-Context Learning (ICL) to generate contextually relevant error instances within candidate fragments. Note that error types vary by financial scenes. For example, in contracts and legal documents, we focus on errors such as clause conflicts, incorrect legal references, and ambiguous expression, while for research reports, which often contain dense numerical and temporal data, we emphasize errors like time inconsistencies, incorrect calculations, and numerical unit errors.

\subsubsection{Error Filtering} 
To ensure the quality of the benchmark, all generated errors undergo a two-stage filtering process. 

\textbf{Model-based Filtering:} Generated errors are re-evaluated by GPT-4o. Specifically, each error-injected sentence replaces its original version within the fragment. Then, the modified fragments, along with their surrounding context, are input into the generator to verify whether the injected errors are contextually appropriate. For calculation errors, we prompt the model to output the corresponding calculation formulas rather than directly judging whether the error is appropriate, as shown in Figure~\ref{fig:cal} in Appendix~\ref{sec:prompt}.

\textbf{Manual Verification:} To eliminate LLMs' knowledge bias, a team of five experts manually verifies and selects the remained errors. This process includes removing errors that may introduce unintended errors and refining error seeds for LLM generation. Besides, the number of error instances per document is controlled to ensure realism. Details of annotation guidelines and consistency are provided in Appendix~\ref{sec:anno}.

\input{bar_plot}

\subsection{Statistics}
FinED-Bench is primarily Chinese-centric, with its statistical information summarized in Figure~\ref{fig:pro_error}, and its distributional comparison with real-world financial documents reported in Table~\ref{tab:comwreal}. To ensure a more balanced and fair evaluation across different error categories, We slightly reduce the proportion of general knowledge errors while increasing the proportion of financial reasoning errors. To further assess the performance of models beyond Chinese, we additionally construct a small English-centric dataset (i.e., FinED-Bench-EN). Detailed statistic and evaluation results on this dataset are presented in Section~\ref{sec:enben}. 
Each document is stored in JSON format as illustrated in Figure~\ref{fig:example} in Appendix~\ref{sec:appa}.

%% file: bar_plot.tex
\begin{table*}[!htbp]
    \centering
    \small
    \begin{tabular}{c|c|c|c|c|c|c|c}
    \toprule
        \multirow{2}{*}{\textbf{Name}} & \multirow{2}{*}{\textbf{\#Docs}} & \multirow{2}{*}{\textbf{Avg. \#Length}}&  \multirow{2}{*}{\textbf{\#Errors}} & \textbf{Avg \#Errors}  & \multicolumn{3}{c}{\textbf{Error Type (\%)}}   \\ \cline{6-8} 
         & & & & \textbf{(Doc)} & \textbf{CKE} & \textbf{FKE} & \textbf{FRE} \\ \midrule
        Real-world Data & 50  & 4,727.8 & 187 & 3.7 & 54.5 & 27.7 & 17.8 \\
         FinED-Bench & 973 & 4,544.7 & 4,123 & 4.2 & 42.9 & 30.3 & 26.8 \\
         FinED-Bench-Hard & 24 & 71,536.0 & 83 & 3.5 & 41.7 & 30.2 & 28.1 \\ 
         \bottomrule
    \end{tabular}
    \caption{Statistical comparison between FinED-Bench and real-world financial documents.}
    \label{tab:comwreal}
\end{table*}

\begin{figure}
    \centering
    \includegraphics[width=0.9\linewidth]{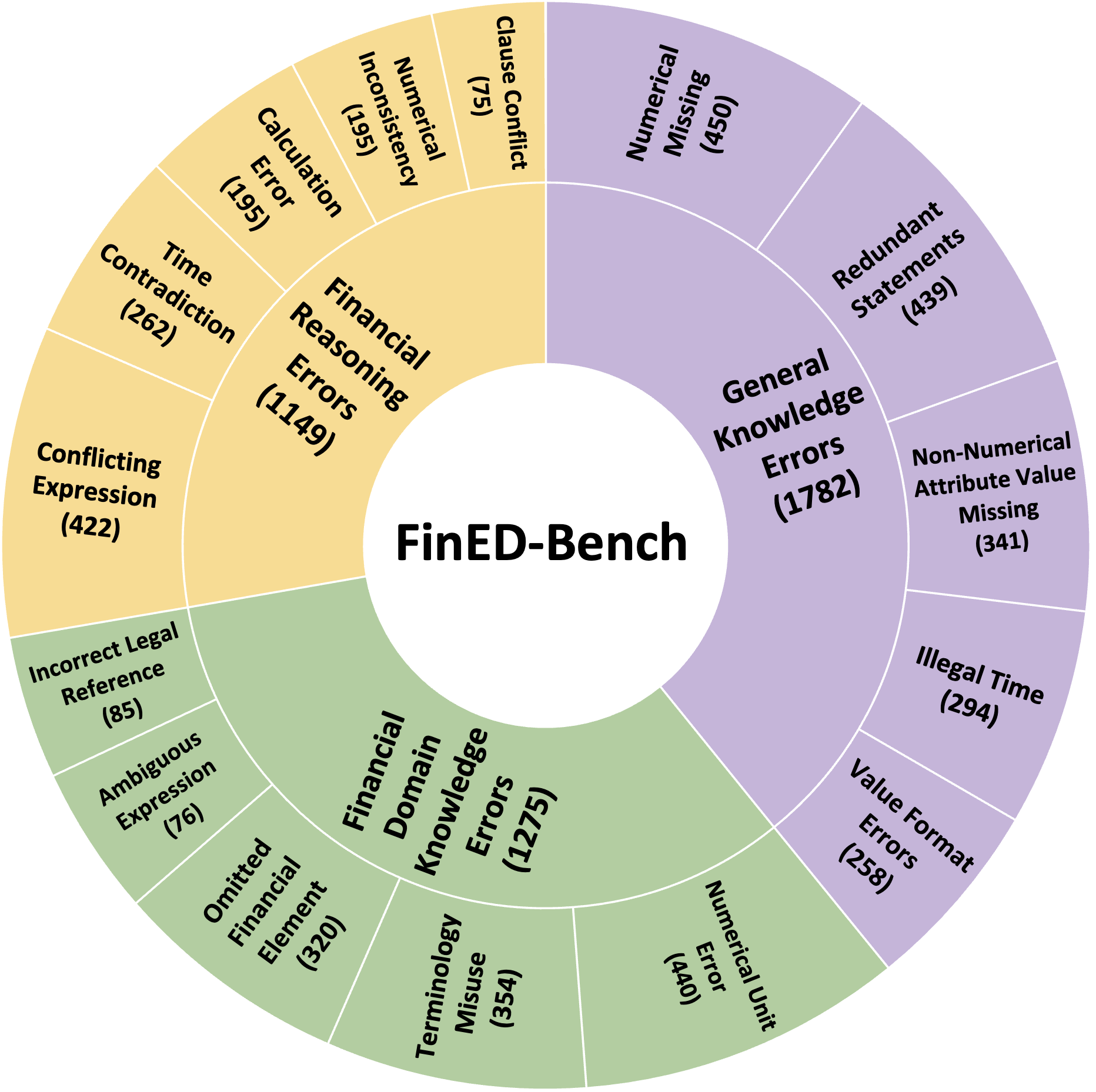}
    \caption{Number of instances per error type in FinED-Bench.}
    \label{fig:pro_error}
\end{figure}

%% file: 04.Experiment.02.tex
\subsection{Overall Setup}
\textbf{LLMs:}
We evaluate FinED-Bench using a diverse set of baseline models spanning both general-purpose and domain-specific LLMs from API-based and open-source domains.
Our evaluation includes models from the Qwen series (Qwen3-8B/14B, Qwen2.5-7B-Instruct, DeepSeek-R1-0528-Qwen3-8B), specialized financial models (Dianjin-R1-7B, Fin-R1), and commercial models (GPT-4o, GPT-4o-mini, GPT-3.5-turbo). Detailed specifications for all baseline models are provided in Appendix~\ref{app:llm-baselines}.


\textbf{Metrics:}
To evaluate the model performance in recognizing financial errors in documents, we use three metrics: \textbf{Precision (Pre.)}, \textbf{Recall (Rec.)}, \textbf{F1-score (F1)}. Specifically, a sentence extracted by LLMs is considered correctly, if: 1) it either exactly matches or contains the standard sentence, and 2) the error type is accurately classified.

\begin{table*}[t]
    \centering
    \small
    \resizebox{\textwidth}{!}{
    \begin{tabular}{l|ccC|ccC|ccC|P }
        \toprule
       \multirow{2}{*}{\textbf{Model}} & \multicolumn{3}{c|}{\textbf{\makecell{General Knowledge \\ Errors}}} & \multicolumn{3}{c|}{\textbf{\makecell{Financial Domain \\ Knowledge Errors}}} & \multicolumn{3}{c|}{\textbf{\makecell{Financial Reasoning \\ Errors}}} &  \textbf{Overall} \\ \cline{2-11}
       & Pre. & Rec. & F1 & Pre. & Rec. & F1 & Pre. & Rec. & F1 & F1 \\ \hline
         \rowcolor{bgcolor} \multicolumn{11}{c}{\textbf{General Large Language Models}} \\ \hline
         Qwen2.5-7B-Instruct & 20.87 & 11.32 & 14.67 & 13.20 & 6.09 & 8.33 & 5.31 & 2.94 & 3.79 &  9.85 \\
         Qwen3-8B (no thinking) & 29.67 & 21.50 & 24.93 & 14.89 & 22.95 & 18.06 & 7.07 & 10.95 & 8.60 & 17.44\\
         Qwen3-8B & \underline{54.21} & 42.43 & 47.60 & \underline{50.52} & 31.22 & 38.60 & \underline{37.33} & 22.36 & 27.96 & 39.99 \\ 
         DeepSeek-R1-0528-Qwen3-8B & \textbf{55.39} & 19.72 & 29.09 & 50.28 & 14.76 & 22.82 & 36.08 & 8.46 & 13.71 & 23.31 \\
         Qwen3-14B (no thinking) & 31.20 & 36.68 & 33.72 & 24.21 & 29.60 & 26.63 & 16.32 & 16.96 & 16.63 & 27.19 \\ 
         Qwen3-14B & 53.15 & \underline{49.77} & \underline{51.41} & \textbf{53.59} & \underline{33.33} & \underline{41.10} & \textbf{40.73} & \underline{22.63} & \underline{29.10}  & \underline{43.15}\\ 
         GPT-3.5-turbo &  20.13 & 10.79 & 14.05 & 5.69 & 2.68 & 3.65 & 10.00 & 1.37 & 2.41 & 7.17\\ 
         GPT-4o-mini & 24.94 & 36.69 & 29.69 & 13.98 & 4.98 & 7.34 & 18.11 & 7.88 & 10.98  & 18.90\\ 
        GPT-4o   & 39.16 & \textbf{78.83} & \textbf{52.33} & 43.23 & \textbf{52.36} & \textbf{47.36} & 28.42 & \textbf{57.35} & \textbf{38.00} & \textbf{48.34} \\ \hline
         
         \rowcolor{bgcolor} \multicolumn{11}{c}{\textbf{Financial Large Language Models}} \\ \hline
         Fin-R1 & 16.62 & 3.39 & 5.63 & 5.04 & 0.97 & 1.63 & 3.49 & 0.55 & 0.95 & 3.19\\ 
         Dianjin-R1-7B & 26.47 & 14.47 & 18.72 & 36.10 & 11.28 & 17.19 & 18.94 & 5.61 & 8.66 & 15.81 \\ \hline
         \rowcolor{bgcolor} \multicolumn{11}{c}{\textbf{Human}}\\\hline
         &  83.33 & 60.61 & 70.18 & 85.00 & 44.74 & 58.62 & 84.62 & 44.00 & 57.89 & 63.63 \\
    \bottomrule
    \end{tabular}}
    \begin{tablenotes}[flushleft]
    \footnotesize
    \item \textit{Note:} Details for the human baseline are provided in Appendix~\ref{app:human}.
    \end{tablenotes}
    \caption{Performance of Different LLMs on FinED-Bench (\%). \textbf{Bold} indicates the best performance and \underline{underlined} indicates the second-best performance within each metric.}
    \label{tab:overall}
    \vspace{-3mm}
\end{table*}

\begin{figure}[t]
    \centering
    \includegraphics[width=\columnwidth]{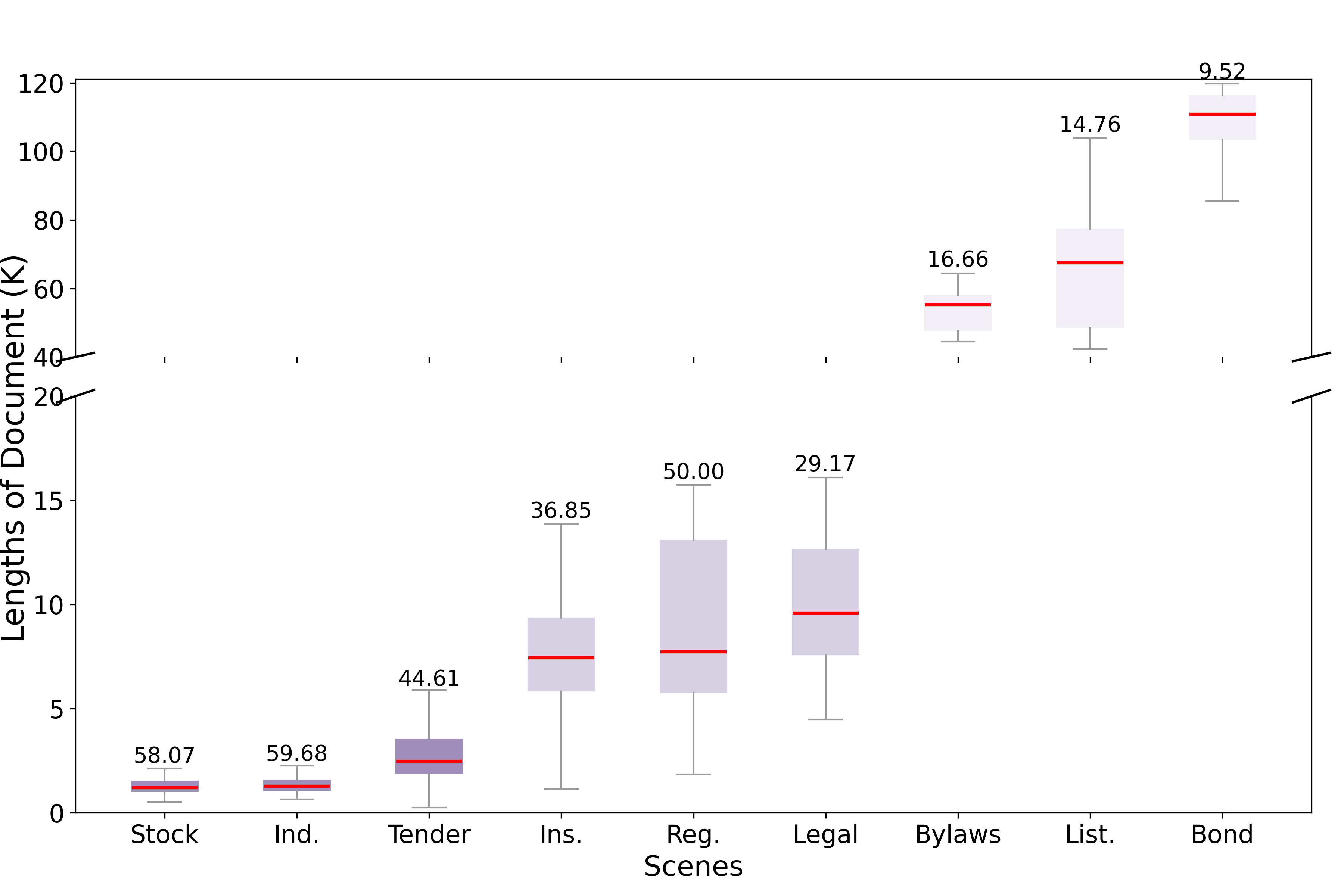}
    \caption{Document Length Distribution and Model Performance (F1) Across Different Scenes (\%). Red lines indicate the median document length for each scene. Black numbers above each box represent the highest F1 scores achieved by the evaluated models.}
    \label{fig:lenf}
\end{figure}

\begin{figure*}[t]
    \centering
    \includegraphics[width=0.9\textwidth]{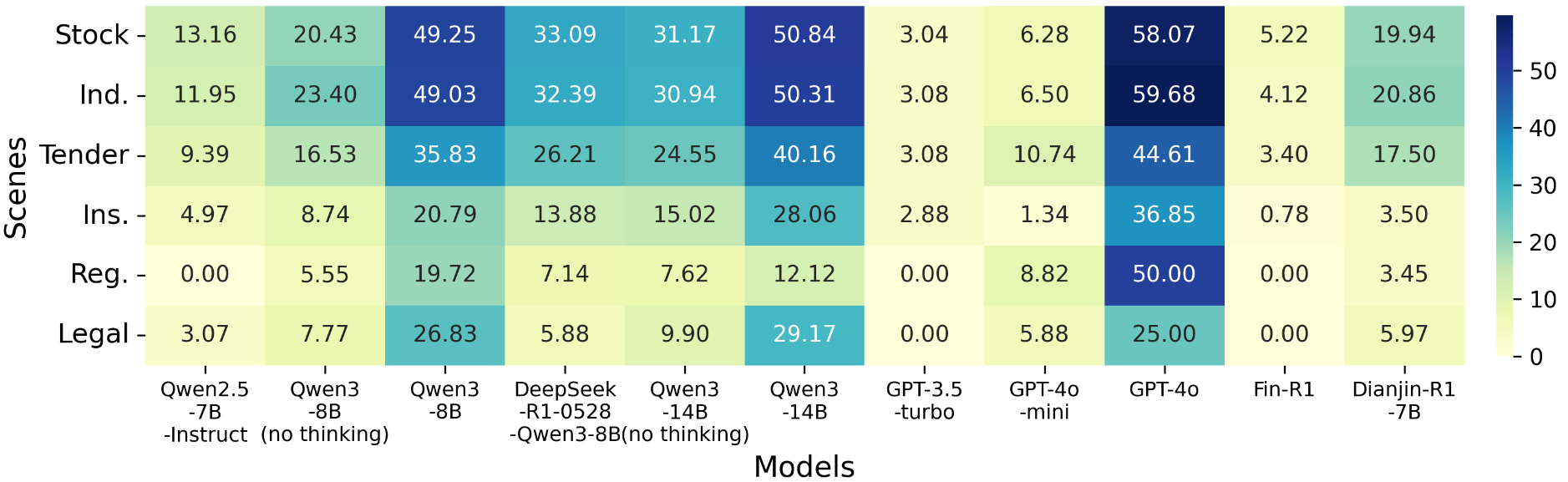}
    \caption{Heatmap of F1-scores (0-100\%) for Each Model (Columns) Across Different Scenes (Rows) in FinED-Bench. Dark blue indicates high performance, while light green indicates low performance.}
    \label{fig:f14scene}
\end{figure*}

\begin{figure*}[t]
    \centering
    \includegraphics[width=0.75\textwidth]{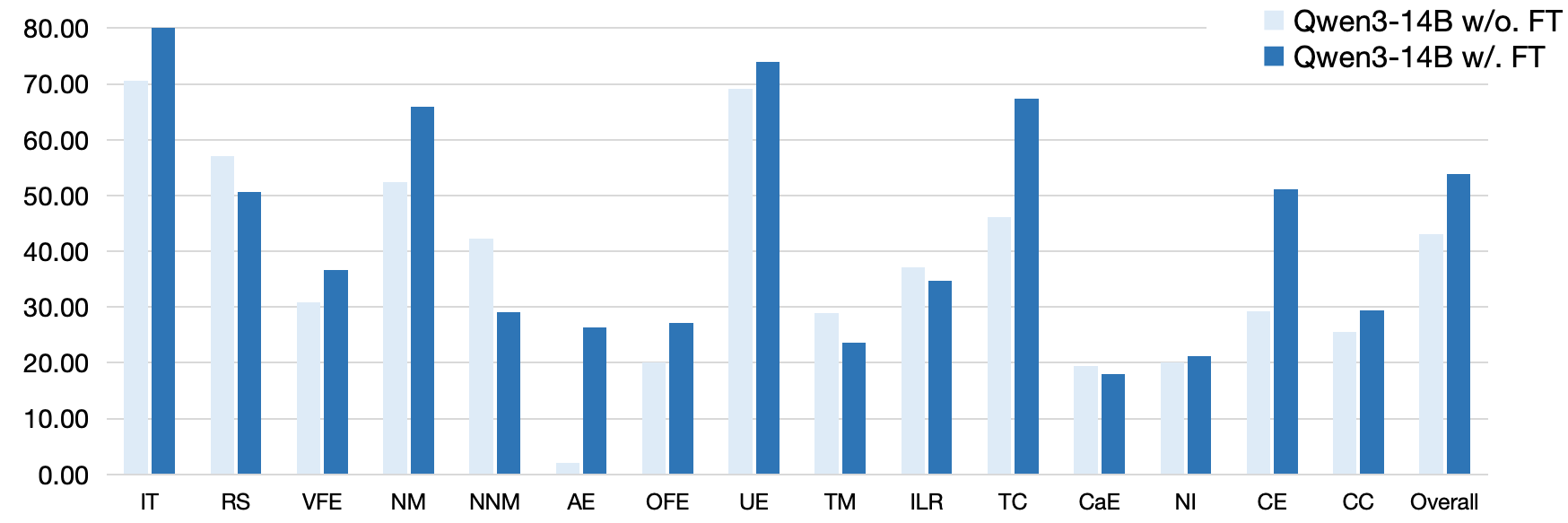}
    \caption{Impact of Supervised Fine-Tuning on Qwen3-14B Across Different Error Types. 
    }
    \label{fig:ft}
    \vspace{-3mm}
    
\end{figure*}

\begin{table}[t]
    \centering
    \small
    \begin{tabular}{c|c|c|c}
        \toprule
            \textbf{Tasks} & \textbf{Metrics} & \textbf{before} & \textbf{after} \\ \midrule
            \multirow{2}{*}{MCQs} & ACC & 73.86 & 73.08 ($\downarrow$ 0.78) \\
            & BLEU-4 & 14.92 & 15.27 ($\uparrow$ 0.35) \\ 
            Fin\_MT &  BLEU-4 & 19.96 & 21.38  ($\uparrow$ 1.42) \\ 
            Fin\_TC & ACC & 66.67 & 68.89 ($\uparrow$ 2.22)  \\
            Fin\_RE & F1 & 47.87 & 45.87($\downarrow$ 2.00 ) \\ 
            Fin\_TG & ROUGE-L & 22.54 & 21.91 ($\downarrow$ 0.63) \\
        \bottomrule
    \end{tabular}
    \caption{Comparison of Performances on Other Financial Tasks from CFLUE before and after Supervised Fine-Tuning. MCQs = multiple-choice questions; Fin\_MT = Financial Translation from English to Chinese; Fin\_TC = Financial Text Classification; Fin\_RE = Financial Relation Extraction; Fin\_TG = Financial Text Generation.}
    \label{tab:general}
\end{table}
\vspace{-4mm}

\subsection{Main Results}
\subsubsection{Overall Performance}

Table~\ref{tab:overall} presents a comparative evaluation of general-purpose and domain-specific LLMs across three categories of financial document errors: General Knowledge, Financial Domain Knowledge, and Financial Reasoning Errors. Details on 15 subtypes of errors are shown in Table~\ref{tab:main_result} in Appendix. Key findings are summarized as follows:

\begin{finding}{1}
Current LLMs struggle with financial reasoning. 
Even GPT-4o show a significant performance drop in this category compare to other categories.
\end{finding}

GPT-4o achieves the best performance on General Knowledge Errors (F1 = 52.33\%) but declines sharply on Financial Reasoning Errors (F1 = 38.00\%). This trend is consistent across other models: Qwen3-14B drops from 51.41\% to 29.10\%, and Qwen3-8B from 47.60\% to 27.96\%. Financial reasoning errors are more challenging for LLMs because they require multi-step calculations and the integration of financial-specific regulations, whereas the other two types of errors rely primarily on commonsense reasoning and factual recall.

\begin{finding}{2}
The performance of domain-specific LLMs is largely constrained by the capabilities of their base models.
\end{finding}

Dianjin-R1-7B, a financial-domain LLM fine-tuned from Qwen2.5-7B-Instruct, achieves a higher overall F1 score than its base model (15.81\% vs. 9.85\%). In contrast, Fin-R1 exhibits a decline in performance, with its overall F1 score dropping from 9.85\% to 3.19\%. Despite these changes, both Dianjin-R1-7B and Fin-R1 remain substantially below the performance of recent general-purpose LLMs such as Qwen3-8B (39.99\%) and Qwen3-14B (43.15\%). These results indicate that (1) the foundational capabilities of base models largely determine the upper bound of performance achievable through domain-specific fine-tuning; (2) fine-tuning on related tasks, such as QA, text summarization, or classification, does not improve the error detection ability of models.

\begin{finding}{3}
GPT-4o shows a high-recall but low-precision pattern on this task, primarily due to its overly sensitive detection.
\end{finding}

Compared with other LLMs, GPT-4o tends to identify a much larger set of candidate errors, which greatly improves its recall but at the cost of precision. As a no-thinking model, GPT-4o struggles to ensure internal inconsistency in its response, leading to numerous false positive and misclassified error types. For example, in ``The scale of waste incineration has reached 975.9 million square meters per day'', the span ``square meters per day'' should be identified as a unit error, but GPT-4o classifies it as a format error; and in ``Recently, Xiangyuan Culture and Tourism released its 2024 annual report and the first-quarter report for 2025'', GPT-4o flags ``2025'' as an illegal time error. 

\begin{finding}{4}
Reasoning capabilities significantly enhance LLM performance on error detection by enabling them to infer implicit relationships within context.
\end{finding}

The reasoning-ablated (``no-thinking'') variants of Qwen3-8B and Qwen3-14B consistently underperform their full counterparts across all error categories. Specifically, Qwen3-8B without reasoning achieves 24.93\% F1 on General Knowledge Errors compared to 47.60\% with reasoning, while Qwen3-14B shows a similar pattern (33.72\% vs. 51.41\%). Distilling DeepSeek-R1-0528~\cite{deepseekai2025deepseekr1incentivizingreasoningcapability} reasoning-chain data into Qwen3-8B improves performance over its no-thinking variant across all categories, yet it still lags behind the full Qwen3-8B, suggesting that fine-tuning may have partially compromised the model’s original capabilities.

\subsubsection{Performance by Scenes}
To investigate the influence of document characteristics on model performance, we compare LLM performance across different financial scenarios. Figure~\ref{fig:lenf} and~\ref{fig:f14scene} present the performance distribution by scenes, highlighting key observations:

\textbf{Longer documents significantly degrade model performance:} As shown in Figure~\ref{fig:lenf}, the performance declines sharply on longer documents. The highest F1 scores occur in Stock Research Reports (58.07\%) and Industry Research Reports (59.68\%), both very short. Performance drops to 44.61\% for Tender Announcements, then further to 38.85\% for Insurance Contracts and 29.17\% for Legal Documents. An exception is Regulatory Documents, which achieve 50.0\%, likely because some regulatory documents in the training data overlap with current regulations, which have changed little. The longest document types perform worst: Company Bylaws (16.66\%), Listing Prospectuses (14.76\%), and Bond Prospectuses (9.52\%). Figure~\ref{fig:f14scene} confirms this trend across models. For example, Qwen3-14B achieves 50.84\% and 50.31\% F1 scores in the two shortest-document scenes, while scores drop substantially for longer documents, as do Qwen3-8B and GPT-4o.  

\subsubsection{Performance of Supervised Fine-tuning}
To enhance model performance on this task, we construct a supervised fine-tuning dataset using the same pipeline as our benchmark, but without human verification. Further details are provided in Appendix~\ref{sec:ft}.

\textbf{Fine-tuning on financial error detection data substantially enhances the models' ability to detect financial errors:} As shown in Figure~\ref{fig:ft}, after fine-tuning on financial error detection data, Qwen3-14B achieves a 10.70\% improvement in overall F1 score (from 43.15\% to 53.85\%). The improvements are particularly notable for challenging error types: Conflicting Expression increases from 29.21\% to 51.06\%, and Ambiguous Expression rises from 2.13\% to 26.42\%. Fine-tuning also enhances performance on format-based errors, with Illegal Time improving from 70.53\% to 80.00\% and Numerical Missing increasing from 52.48\% to 65.99\%. Even complex reasoning errors show gains, with Time Contradiction improving from 46.07\% to 67.32\% and Clause Conflict rising from 25.53\% to 29.36\%.

\textbf{Fine-tuning preserves the model's generalization ability:} We further evaluate the fine-tuned model’s generalization through two dimensions: knowledge and application, with five financial tasks from CFLUE~\cite{zhu2024cflue}: MCQs, Fin\_MT, Fin\_TC, Fin\_RE, and Fin\_TG. MCQs represent the knowledge dimension, while the remaining four tasks assess application-oriented capabilities. The results in Table~\ref{tab:general}  show that the fine-tuned model maintains comparable performance across these tasks, and in several cases even shows slight improvements over the original model.

%% file: 05.Conclusion.01.tex
In this paper, we present FinED-Bench, a benchmark specifically designed to assess the error detection capabilities of LLMs within financial documents. To avoid data contamination, we also propose a semi-automatic construction pipeline that involves crawling raw financial documents from the internet and injecting domain-specific errors. The experimental results show that, while incorporating reasoning significantly enhances model performance, even recent LLMs, like GPT-4o and Qwen3-14B, still struggle with detecting errors in financial documents. Furthermore, we observe that existing domain-specific LLMs, which are often fine-tuned for specific downstream tasks, such as stock movement prediction, demonstrate limited improvements in the error detection tasks, compared with their base models.